%% file: main.tex
\documentclass[10pt,twocolumn,letterpaper]{article}

\usepackage{cvpr}              
\usepackage{algorithm}
\usepackage{amssymb}
\usepackage{amsmath}
\usepackage{siunitx} 
\input{preamble}

\definecolor{cvprblue}{rgb}{0.21,0.49,0.74}
\usepackage[pagebackref,breaklinks,colorlinks,allcolors=cvprblue]{hyperref}

\def\paperID{} 
\def\confName{CVPR}
\def\confYear{2026}
\def\fname{Random Parameter Pruning Attack}
\def\name{RaPA}
\title{RaPA: Enhancing Transferable Targeted Attacks via \\ Random Parameter Pruning}

\author{
    Tongrui Su \quad 
    Qingbin Li \quad 
    Shengyu Zhu\thanks{Corresponding author.} \quad 
    Wei Chen\footnotemark[1] \quad 
    Xueqi Cheng \\
    State Key Lab of AI Safety, Institute of Computing Technology, Chinese Academy of Sciences \\
    University of Chinese Academy of Sciences \\
    \texttt{\{sutongrui25s,\,liqingbin24z,\,zhushengyu,\,chenwei2022,\,cxq\}@ict.ac.cn}
}

\begin{document}
\maketitle

\input{sec/0_abstract}

\input{sec/1_introduction}
\input{sec/3_preliminary}

\input{sec/4_method}

\input{sec/5_experiment}
\input{sec/acknowledge}
{
    \small
    \bibliographystyle{ieeenat_fullname}
    \bibliography{main}
}

\input{sec/X_supplementary}

\end{document}

%% file: preamble.tex
\DeclareMathOperator*{\argmax}{arg\,max}

\usepackage{algorithmicx}
\usepackage{algpseudocode}
\usepackage{amssymb}
\usepackage{amsmath}
\usepackage{siunitx}

\usepackage{booktabs}

\usepackage{multirow}
\PassOptionsToPackage{table}{xcolor}  
\usepackage{xcolor}
\usepackage{colortbl}

\definecolor{Gray}{gray}{0.92}
\usepackage{subcaption}
\DeclareCaptionLabelFormat{offset}{\hspace*{0.5cm}(#2)}

%% file: sec/0_abstract.tex
\def\fname{Random Parameter Pruning Attack}
\def\name{RaPA}

\begin{abstract}

Compared to untargeted attacks, targeted transfer-based attack still suffers from much lower Attack Success Rates (ASRs),  although significant improvements have been achieved by kinds of methods, such as diversifying input, stabilizing the gradient, and re-training surrogate models. In this paper, we find that  adversarial examples generated by existing methods rely heavily on a small subset of surrogate model parameters, which  limits their transferability to unseen target models. Inspired by this finding, we propose Random Parameter Pruning Attack (RaPA), which introduces parameter-level randomization during the attack process. At each optimization step, RaPA randomly prunes model parameters to generate diverse yet semantically consistent surrogate variants. We show that this parameter-level randomization is equivalent to adding an importance-equalization regularizer, thereby alleviating the over-reliance issue. Extensive experiments across both CNN and Transformer architectures demonstrate that RaPA substantially enhances transferability. In the challenging case of transferring from CNN-based to Transformer-based models, RaPA achieves up to 11.7\% higher average ASRs than state-of-the-art baselines (with 33.3\% ASRs), while being training-free, cross-architecture efficient, and easily integrated into existing attack frameworks. Code is available on https://github.com/molarsu/RaPA.

\end{abstract}

%% file: sec/1_introduction.tex
\section{Introduction} \label{intro}
Deep neural networks have become prevalent in computer vision applications \cite{ResNet, DenseNet, ViT}, but are  highly vulnerable to maliciously crafted inputs, called adversarial examples \cite{First, FGSM}. A major concern is their transferability, that is, adversarial examples generated using a white-box model can directly fool other black-box models, without any access to their architectures, parameters or gradients \cite{TransferSurvey}. Since this type of attacks, usually referred  to as transfer-based attacks, do not require any interaction with the target model, they pose severe security risks to real-world machine learning systems. Therefore, studying effective transfer-based attack methods is crucial 
to understand the vulnerabilities and further enhance model robustness.

This paper focuses on  targeted transfer-based attacks with a \emph{single surrogate model}, where the goal is to deceive black-box models to classify input images into a specific incorrect category. Due to the high complexity of decision boundaries,  existing methods still have noticeably lower Attack Success Rates (ASRs) in the targeted setting than in the untargeted \cite{Logit, CFM}. 
A key observation is that the generated adversarial examples tend to overfit the  surrogate model but fail to generalize  to other models. To improve transferability, various strategies have  been proposed.  Observing that multiple surrogate models can help  enhance transferability but in practice finding proper models for the same task is not easy \cite{MI,Liu2016DelvingIT,liu2024scaling,TRS}, model self-ensemble \cite{T-sea,MUP,Ghost,SETR}  tries to create multiple models from  an accessible model. Input transformation   \cite{DI,RDI,ODI,Admix,BSR,SIA,CFM, FTM} applies different transformations to  inputs and diversify input patterns  to reduce  overfitting.  A notable method is  Clean Feature Mixup (CFM) \cite{CFM}, which randomly mixes high-level  features with shuffled clean features. Building upon it, Feature Tuning Mixup (FTM) \cite{FTM}  introduces learnable and attack-specific feature perturbations, achieving new state-of-the-art performance in transferability. Despite these progresses, there is still much room for further improvement.

In this work, we take a different perspective and identify a previously overlooked cause of the poor transferability: the generated adversarial perturbations rely excessively on a small subset of parameters in the surrogate model, which limits their generalization to other models that have different parameter configurations. In other words, adversarial perturbation in existing methods tends to exploit a few “shortcut” parameters, leading to strong white-box performance but poor black-box transferability.

To mitigate this issue, we propose Random Parameter Pruning Attack (RaPA)  that introduces parameter-level randomization into the attack process. At each optimization step, RaPA randomly prunes a subset of parameters in the surrogate model and uses multiple masked variants to update the adversarial example. We show that taking the expectation over such random masks is equivalent to adding an importance regularization term that aims to  equalize parameter contributions, thus preventing over-reliance on a few dominant parameters.
Conceptually, RaPA can be viewed as a self-ensemble method: each randomly pruned model represents a diverse yet semantically consistent variant of the surrogate. Previous self-ensemble approaches SASD-WS \cite{SASD-WS}, MUP \cite{MUP}, and Ghost Network \cite{Ghost} rely on training-based model enhancement, deterministic pruning metrics, and  structural perturbations, respectively. In contrast, RaPA is training-free, cross-architecture efficient, and straightforward to implement.

\begin{figure}[t]
    \centering
    \includegraphics[width=0.95\linewidth]{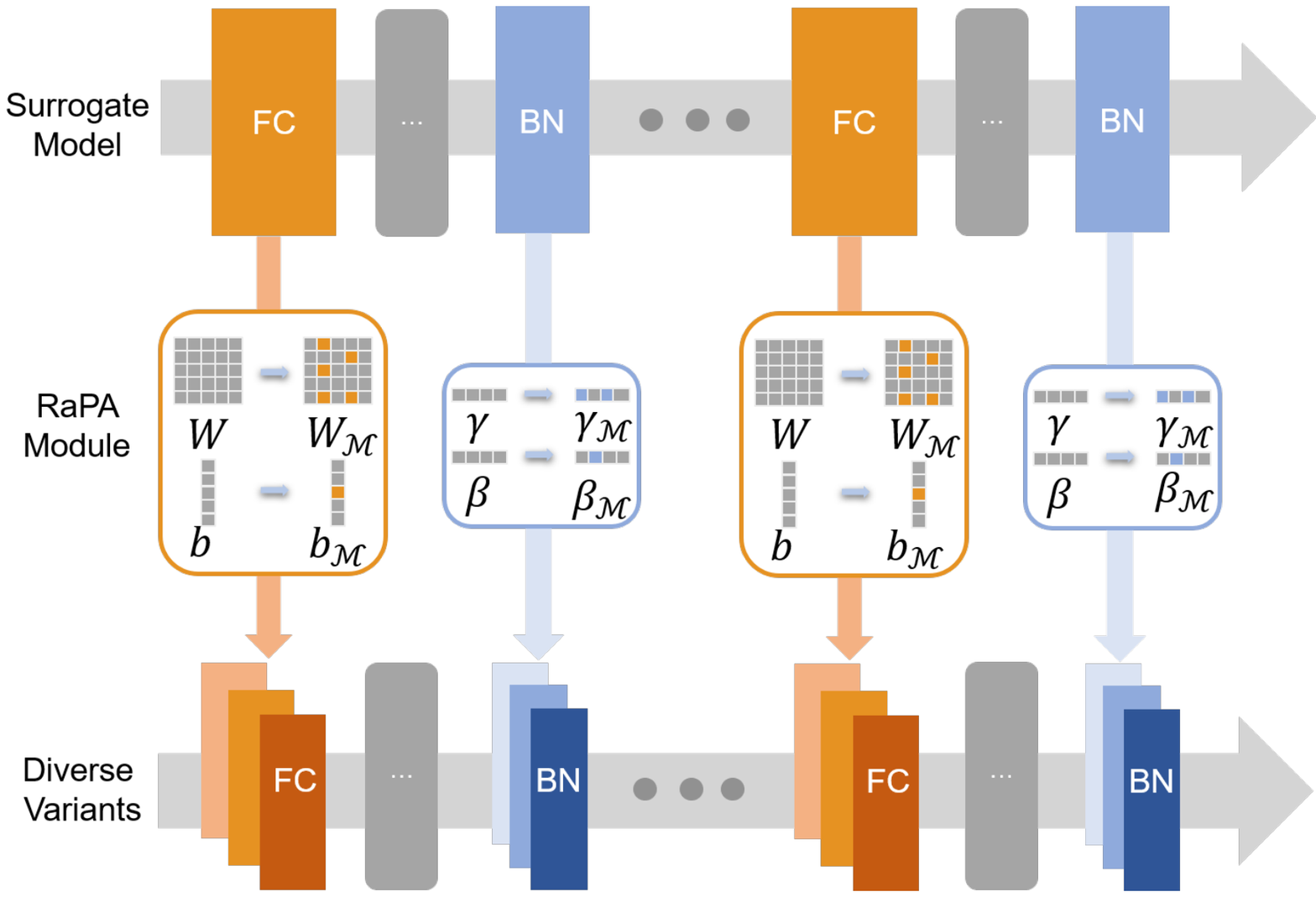}
         \vspace{-0.5em}
    \caption{
        An illustration of the proposed method \name. We apply \name\ to selected layers in the surrogate model to create multiple and diverse variants  at each iteration.  
}
     \vspace{-0.5em}
    \label{fig:overview}
    \vspace{-1 em}
\end{figure}

We evaluate the proposed method 
across various CNN- and  Transformer-based target models, as well as against several defense methods. The experimental results show that RaPA outperforms other state-of-the-art methods. In particular, in the challenging scenario of transferring from CNN-based model to Transformer-based models, \name\ achieves 11.7\% and 17.5\% higher average ASRs with ResNet-50 \cite{ResNet} and DenseNet-121 \cite{DenseNet} as surrogate models, respectively. {Moreover,  \name~achieves the highest performance gain when scaling the compute for crafting adversarial examples. Specifically, with ResNet-50 as surrogate model, increasing the optimization iterations from 300 to 500 and   number of forward-backward passes per iteration from 1 to 5 boosts the average ASR by 15.9\%.}

To summarize, our main contributions are  as follows:
\begin{itemize}

   \item We show that adversarial examples from existing transfer-based attacks rely heavily on a tiny subset of parameters in the surrogate model. Alleviating this over-reliance can in turn enhance the transferability of attack.
   
   \item We propose the RaPA, which introduces parameter-level randomization during attack optimization. We show, both intuitively and empirically, that random pruning implicitly equalizes parameter importance, acting as a regularizer to mitigate the over-reliance issue.
   
   \item Experiments across diverse surrogate and target models demonstrate that RaPA consistently outperforms existing methods. \name\ further benefits from increased computational budget, achieving larger improvements when scaling optimization iterations or inference steps.

\end{itemize}

%% file: sec/3_preliminary.tex
\input{tables/pilot_study}

\section{Preliminary}
This section introduces the background of adversarial attacks and briefly reviews related works on targeted transfer-based attacks. See \Cref{app:related_work} for more related work.

\label{sec:pre}
\subsection{Background}
Consider a classification task where the model is defined as a function $f: \mathbb R^n \to\mathcal{Y}$ that maps an input $x\in\mathbb R^n$ to a label in the set $\mathcal{Y}$ consisting of all the labels. Given a clean image~$x$ with its true label $y\in\mathcal {Y}$,  untargeted attacks aim to find an adversarial example $x_{\mathrm{adv}}\in\mathbb{R}^n$ that is similar to $x$ but misleads the model to produce an incorrect prediction, 
i.e. $f(x_\mathrm{adv}) \neq y$. Here  `similarity' is usually measured by an $\ell_p$-norm, e.g., $\|x_{\mathrm{adv}}-x\|_p\leq\epsilon$ where $\epsilon>0$ is a predefined perturbation budget. For targeted attacks, the goal is to modify the model prediction to a particular target label $y_\mathrm{tar}$, that is, $f(x_\mathrm{adv})=y_\mathrm{tar}\neq y$. In this work, we will focus on targeted attacks. 

In the white-box setting where the model is fully accessible, adversarial example can be obtained by the following:
\begin{align}
\label{eqn:objective}
\argmax_{x_{\mathrm{adv}}}~\mathcal{L}\left(f(x_\mathrm{adv})\right),~~ \text{s.t.}~~\|x_\mathrm{adv}-x\|_p \leq \epsilon,
\end{align} 
where \( \mathcal{L}(\cdot)\) is a loss function (e.g., cross-entropy loss). The Fast Gradient Sign Method (FGSM) \cite{FGSM} uses the gradient direction to solve this problem and craft adversarial examples, while Iterative FGSM (I-FGSM) \cite{I-FGSM} extends this idea to an iterative scheme. In particular, at each iteration $t$, adversarial example  is updated by adding a small perturbation:
\begin{align}
\label{eqn:IFGSM}
    x_{\mathrm{adv}}^t =  x_{\mathrm{adv}}^{t-1} + \alpha\cdot\mathrm{sign}\left(\nabla_{x_{\mathrm{adv}}^{t-1}} \mathcal{L}\left(f(x_{\mathrm{adv}}^{t-1})\right)\right),
\end{align}
where $x_{\mathrm{adv}}^0=x$ and $\alpha>0$ is a step size. To make the generated adversarial example satisfy the perturbation budget constraint, a straightforward way is to  project $x_{\mathrm{adv}}^t$ into the $\epsilon$-ball of $x$.

\subsection{Related Work}
In  black-box settings, the gradient information is not available and we only have limited access to the target model. 
Transfer-based method assumes that the adversarial examples generated on one model to mislead not only that model but also other models \cite{First}. 

To improve  transferability, many  methods have been proposed.
The first class, input-transformation techniques, applies a transformation $\mathcal T$ and uses $\nabla_{x_{\mathrm{adv}}^{t-1}} \mathcal{L}\left(f(\mathcal T(x_{\mathrm{adv}}^{t-1}))\right)$ as the gradient.
Diverse Inputs (DI) \cite{DI} and its variant Resized DI (RDI) \cite{RDI} apply random transformations to increase input variation during optimization.
Translation-Invariant (TI) \cite{TI} averages gradients over translated inputs to reduce location sensitivity. Structure Invariant Attack (SIA) \cite{SIA} and Block Shuffle and Rotate (BSR) \cite{BSR} both perform block-level local transformations, with SIA applying diverse transforms and BSR focusing shuffle and rotate operations. Object-based DI (ODI) \cite{ODI} generates adversarial examples rendered on 3D objects, while Admix \cite{Admix} mixes inputs with random samples from other classes. CFM \cite{CFM} extends it to the feature space with competing noises, while FTM \cite{FTM} further adds learnable, attack-specific perturbations to achieve state-of-the-art transferability.

The second class focuses on stabilizing gradient updates to improve transferability. Momentum Iterative FGSM (MI-FGSM) \cite{MI} incorporates a momentum term into I-FGSM to help avoid local optima.
Scale-Invariant (SI) optimization \cite{SI} improves transferability by applying perturbations across multiple scaled copies of the input, leveraging the scale-invariance property of deep models.

Beyond the above  methods, another type of approaches re-train the surrogate model to enhance transferability, e.g., DSM \cite{DSM} and SASD-WS \cite{SASD-WS} improve model generalization and transferability through knowledge distillation or sharpness-aware self-distillation.

Closely related to the present work is self-ensemble \cite{T-sea,MUP,Ghost,SETR}, which creates
multiple models from only one surrogate model. 
The self-ensemble method in  \cite{SETR} specifically considers vision Transformer as surrogate model and is denoted as SE-ViT in this paper. Ghost Network \cite{Ghost} perturbs surrogate model to create a set of new models and then samples one model from the set at each iteration. Masking Unimportant Parameters (MUP) \cite{MUP} drops out unimportant parameters according to a predefined Taylor expansion-based metric, while Diversity Weight Pruning(DWP) \cite{DWP} only prunes the parameters with small absolute values.

However, the inherent differences w.r.t.~model architecture, parameter setting, and training procedure between the surrogate and target models still limit the effectiveness of transfer-based methods on certain models and datasets. In the next section, we  show that there is a key aspect that renders the current over-fitting issue of transfer-based methods.

%% file: tables/pilot_study.tex

\begin{table*}[ht]
\centering
\renewcommand{\arraystretch}{1.0}
\resizebox{0.65\textwidth}{!}{
\begin{tabular}{l c c c c c c c c}
\toprule[0.15em]
\textbf{Method} & DI & RDI & SI  & Admix & ODI & BSR & CFM & \name \\
\midrule
\textbf{No pruning } & 98.2 & 98.7 & 98.7 & 98.1 & 98.9 & 98.5 & 98.0 & 98.2 \\

\textbf{Pruning bottom 0.5\%   } & 98.2 & 98.7 & 98.7 & 98.0 & 98.9 & 98.5 & 98.0 & 98.0 \\

\textbf{Pruning top 0.5\%  } & 16.0 & 28.6 & 31.9 & 29.9 & 19.9 & 36.7 &  51.3 & 64.5 \\
\bottomrule[0.15em]
\end{tabular}%
}
\vspace{-0.3em}
\caption{ASRs (\%) before and after pruning the selected subsets of parameters in the surrogate model on the ImageNet-compatible dataset. Detailed experimental setting can be found in \Cref{sec:expsetting}.} 
\label{tab:pilot_study}
\vspace{-0.7em}
\end{table*}

%% file: sec/4_method.tex
\section{Method}
\label{sec:method}
In this section, we first conduct a pilot study to show a key aspect of the overfitting issue in existing transfer-based methods and then propose a random masking based approach. Comparison with related methods is also discussed.

\subsection{Motivation}
We observe that the adversarial perturbations generated by solving Problem~\eqref{eqn:objective} tend to rely heavily on a small subset of parameters in the surrogate model. These parameters may stem from specific training schemes, datasets, or architectural choices. As a result, adversarial examples that strongly depend on these parameters often fail to generalize and mislead other models. Even with state-of-the-art transfer-based attack methods, this issue remains a key factor that can lead to the failure of adversarial example transfer.

To quantify the phenomenon, we conduct a pilot study using the framework of Optimal Brain Damage (OBD) \cite{lecun1989optimal, molchanov2019importance}, which quantifies the importance of each model parameter from the perspective of sensitivity analysis. Specifically, given an adversarial example $x_\mathrm{adv}$ and a loss function $\mathcal{L}(\cdot)$. Let ${\theta}$ represent the entire set of model parameters. The importance of 
a parameter $\theta_i$ in the surrogate model $f$ is computed as:
\begin{align}
    \mathcal{I}(\theta_i) = \frac{\partial^2 \mathcal{L}\left(f(x_\mathrm{adv})\right)}{\partial \theta_i^2} \times \theta_i^2.
    \label{eqn:importance}
\end{align}
This metric reflects how much the loss would change if a parameter $\theta_i$ were removed, and can serve as a proxy for its contribution to the effectiveness of the adversarial example. 

Next, we consider pruning two distinct subsets of the surrogate model’s parameters based on this importance metric: the top 0.5\% most important and the bottom 0.5\% least important parameters(see \Cref{sec:prune_method} for details). For each adversarial example, we instantiate the model, prune the selected subset, and evaluate the ASR on the resulting  model. Table \ref{tab:pilot_study} reports the  ASRs after pruning the two subsets of model parameters. Here we use ResNet50 as the surrogate model and  detailed  setting can be found in \Cref{sec:expsetting}.

We observe that pruning the most important parameters leads to a drastic drop in ASR—more than 46\%, whereas pruning the least important parameters yields negligible impact. This observation suggests that adversarial examples generated by existing methods are highly dependent on the most important parameters, validating our observation on the over-reliance issue. As such, how to further alleviate this strong dependence on specific parameters would be a key to improving the transferability of adversarial examples over existing transfer-based attack methods.

\subsection{Alleviating Over-reliance via Random Parameter Pruning}

As per the pilot study, a direct approach to improving transferability would be to mask the most important parameters at each optimization step, thereby mitigating the over-dependency on them. However, accurately identifying important parameters requires computing second-order derivatives, which is computationally expensive for all parameters.  Although we can approximate them with first-order terms, masking the most important parameters typically causes the surrogate model’s capacity to degrade rapidly, and the resulting adversarial examples may fail to fool the target model—and even the original surrogate model itself (See \Cref{app:mask_most_parameter} for further explanations). To address this problem, we propose to apply random parameter pruning to the surrogate model at each optimize step. This approach avoids expensive computations while achieving the goal of reducing over-reliance on specific parameters.

\vspace{-1em}
\paragraph{Intuition and Theoretical Explanation}
Our core idea is that randomly pruning parameters at different optimization steps encourages the generated adversarial examples to be less dependent on particular parameter subsets. This in turn improves transferability across different target models.
 

We define a random binary mask $\mathcal{M} \in \{0,1\}^{|\theta|}$, where each entry is independently sampled from a Bernoulli distribution: $\mathcal{M}_i \sim \text{Bernoulli}(1 - p)$. Here $p \in [0,1]$ is the probability of masking a parameter. With a small $p$, we would have $\mathbb E[\mathcal{M}_i] \approx 1$. Then the parameter of the model used in the forward pass becomes $\mathcal{M} \odot \theta$, where $\odot$ denotes element-wise multiplication.

Under this setup, the expected loss over random masks can be approximated using a second-order Taylor expansion:
\begin{align}
&~\mathbb{E}_{\mathcal{M}}[\mathcal{L}(f(x_{\mathrm{adv}}; \mathcal{M} \odot \theta))] \nonumber \\ \approx &~\mathcal{L}(f(x_{\mathrm{adv}}; \theta)) 
+ \frac{p(1-p)}{2}\sum_i \frac{\partial^2 \mathcal{L}f(x_\mathrm{adv};\theta)}{\partial \theta_i^2} \theta_i^2,
\label{eqn:taylor_mask}
\end{align}
which is sum of the original loss plus an importance penalty. Minimizing this objective while resampling the mask at each step would force the adversarial example to distribute the importance  over all parameters, making it more robust to different parameters and thus more transferable.

\paragraph{Practical Implementation with DropConnect
}
The above random parameter pruning method is similar to DropConnect method \cite{DropConnect} in training neural networks. We notice that DropConnect is mainly effective in terms of linear layers . We thus apply DropConnect to the weight and bias parameters of linear layers as well as the transformation parameters of normalization layers.  Both types of layers are widely used in mainstream architectures including Transformer \cite{Attention}. Empirically, our ablation study in \Cref{sec:exp} validates the effectiveness of this choice,  compared with convolutional layers. 

We now present our attack method,  \fname~(\name), as summarized in \Cref{alg:1}. Take  linear layer for example.  We perform independent random masking onto the weight and bias (if present)  parameters using Bernoulli sampling. Specifically, for surrogate model $f$, let $W \in \mathbb{R}^{d_{\text{in}} \times d_{\text{out}}}$ denote the weight matrix and $b \in \mathbb{R}^{d_{\text{out}}}$ the bias vector associated with a  linear layer. Here $d_{\text{in}}$ and $d_{\text{out}}$ represent the input and output dimensions, respectively. The corresponding masks for the weight matrix and bias are 
\begin{align} 
    \mathcal{M}_w \sim \text{Bernoulli}(1-p_w), \mathcal{M}_b \sim \text{Bernoulli}(1-p_b), 
    \label{eq:dropconnect_mask} 
\end{align} 
where $\mathcal{M}_w \in \{0, 1\}^{d_{\text{in}} \times d_{\text{out}}}$ and $\mathcal{M}_b \in \{0,1\}^{d_{\text{out}}}$ are the random masks, and  $p_w, p_b \in [0,1]$ are DropConnect probabilities. Then the masked parameters are  computed as
\begin{equation} 
    W_{\mathcal{M}} = \mathcal{M}_w \odot W,~~ b_{\mathcal{M}} = \mathcal{M}_b \odot b, 
    \label{eq:dropconnect} 
\end{equation} 
where $\odot$ denotes the element-wise multiplication. For normalization layer, the same operation is applied similarly to the transformation parameters. The random masks are sampled for each selected layer.

The random masks $\mathcal{M}_w$ and $\mathcal{M}_b$ in Eq.~\eqref{eq:dropconnect_mask} are re-generated for each inference, producing diverse variants of the surrogate model with different parameters if we conduct multiple inferences at an iteration.  In addition, \name\  can be naturally integrated with existing input transformation  and gradient stabilization methods for crafting adversarial examples, as shown in Lines 5 and  8 in \Cref{alg:1}.

\input{tables/alg}

\input{tables/gini}
\vspace{-1em}
\paragraph{Analyzing Parameter Importance with Gini Coefficient}
\label{sec:gini_analysis}

To further verify that our random parameter pruning strategy indeed mitigates the over-reliance on a few dominant parameters, we employ the  Gini coefficient  to measure the distribution of parameter importance across layers. A lower Gini value indicates a more uniform distribution of importance, implying that the adversarial perturbation depends less on specific parameters and generalizes better to unseen models.
The formal definition and detailed computation process of the Gini coefficient are provided in \cref{app:detailed_gini}.

We compute the Gini coefficients based on the parameter importance values $\mathcal{I}(\theta_i)$ defined in Eq.~\eqref{eqn:importance}. The overall and layer-wise results are summarized in Table~\ref{tab:gini}. As shown in Table~\ref{tab:gini}, RaPA achieves the lowest Gini coefficients among all compared methods, suggesting that it effectively flattens the importance distribution and suppresses the over-concentration of sensitivity on a few parameters. This balanced importance allocation leads to improved robustness and transferability across different architectures.

\subsection{Discussion}
\label{sec:discussion}

In this section, we compare \name\ with related self-ensemble methods in more details.

\name\ was proposed to reduce over-dependence on specific parameters, and it turns out to be a self-ensemble method that constructs multiple new models at each iteration.
Existing self-ensemble method \cite{T-sea} targets object detection task, which is different from ours. SE-ViT \cite{SETR} is specifically designed for vision Transformer surrogate models; as shown  in \Cref{sec:exp}, its ASR is  lower than \name\ even when using ViT as surrogate model.  More closely related are Ghost Network \cite{Ghost}  and  MUP \cite{MUP}, but are much outperformed by \name\  (c.f.~\Cref{app:mask_most_parameter} and \Cref{sec:mainresult}).

We now analyze the effectiveness of \name\ from  model ensemble perspective. It has been
hypothesized that an adversarial image that remains adversarial for multiple models is  more likely to transfer to other models \cite{Liu2016DelvingIT}.  \name\ generates independent random masks for each selected layer and also at each optimization iteration. In this sense, it brings in more randomness and further diversification than Ghost Network, MUP and DWP. Specifically, Ghost Network perturbs only skip connections for residual networks (like ResNet-50), while MUP and DWP mask unimportant parameters according to a predefined metricor at each iteration. This observation is also in accordance with  \cite{T-sea,Chen2023RethinkingME,liu2024scaling}, which show that increasing the number of surrogate models generally enhances the transfer attack performance. On the other hand,  from the perspective of  ensemble techniques in machine learning, each variant model should also be  informative about or useful to the targeted task  of image classification \cite{T-sea,Chen2023RethinkingME}. As empirically shown in \Cref{sec:div_util}, \name\ achieves a  good tradeoff in terms of model diversity and utility, thereby enhancing the  attack performance.

%% file: tables/alg.tex
\begin{algorithm}[t!]
\algnewcommand\algorithmicinput{\textbf{Input:}}
\algnewcommand\Input{\item[\algorithmicinput]}
\algnewcommand\algorithmicoutput{\textbf{Output:}}
\algnewcommand\Output{\item[\algorithmicoutput]}
\caption{\fname (\name)}
\label{alg:1}
\begin{algorithmic}[1]
    \Input{Classifier $f$; clean image $x$;  loss function $\mathcal{L}(\cdot)$; max iterations $T$; $\ell_p$ bound $\epsilon$; number of inferences per iteration $S$; DropConnect probabilities $p_w$, $p_b$;  linear and normalization layers $\mathbb{L}$; input transformation $\mathcal{T}$.}
    \Output{Adversarial example $x_{\mathrm{adv}}$}
    \State $x_{\mathrm{adv}}^0\leftarrow x$
    \For{$t = 1 \to T$} 
        \For{$s = 1 \to S$}
            \State Obtain modified model $f_{\mathcal{M}}$ by applying \name\ to each layer in $\mathbb{L}$ according to Eqs.~\eqref{eq:dropconnect_mask} and \eqref{eq:dropconnect}.
            \State $g_s^t\leftarrow\nabla_{x_{\mathrm{adv}}^{t-1}} \mathcal{L}\left(f_{\mathcal{M}}(\mathcal T(x_{\mathrm{adv}}^{t-1}))\right)$
        \EndFor
        \State $g^t\leftarrow \frac{1}{S} \sum g_s^t$
        \State Update $x_{\mathrm{adv}}^t$  with gradient $g^t$ using iterative methods (like MI-FGSM~\cite{MI}).
        \State Project ${x_{\mathrm{adv}}^{t}}$ into the $\epsilon$-ball of $x$.
    \EndFor
    \State \Return $x_{\mathrm{adv}}^T$
\end{algorithmic}
\end{algorithm}

%% file: tables/gini.tex
\begin{table*}[ht]
\centering
\renewcommand{\arraystretch}{1.0}
\small
\resizebox{0.65\textwidth}{!}{
\begin{tabular}{l c c c c c c c c}
\toprule[0.15em]
\textbf{Method} & DI & RDI & SI  & Admix & ODI & BSR & CFM & \name \\
\midrule
\textbf{All Layer Average } & 0.32 & 0.30 & 0.21 & 0.12 & 0.33 & 0.25 & 0.19 & \textbf{0.08} \\
\textbf{Conv Layer } & 0.11 & 0.07 & 0.03 & 0.01 & 0.12 & 0.05 & 0.03 & \textbf{0.00} \\
\textbf{Norm Layer } & 0.51 & 0.52 & 0.37 & 0.22 & 0.53 & 0.44 & 0.34 & \textbf{0.15} \\
\textbf{Linear Layer } & 1.00 & 0.86 & 0.59 & 0.18 & 1.00 & 0.75 & 0.55 & \textbf{0.13} \\
\bottomrule[0.15em]
\end{tabular}%
}
\vspace{-0.3em}
\caption{Gini coefficients of parameter importance across different layers and methods. Lower values correspond to more uniform parameter importance.} 
\label{tab:gini}
\vspace{-0.7em}
\end{table*}

%% file: sec/5_experiment.tex
\section{Experiments}

\label{sec:exp}

\input{tables/targeted_vit_results}

This section empirically validates the effectiveness of our method, using both CNN- and Transformer-based models.

\subsection{Experimental Settings}
\label{sec:expsetting}

\paragraph{Dataset} We utilize the ImageNet-compatible dataset \cite{nips_dataset}, served as the official dataset for the NIPS~2017 Attack Challenge. This dataset contains both ground-truth and targeted labels, making it well-suited for targeted-attack.

\vspace{-1em}

\paragraph{General Setting} We adopt the $\ell_\infty$-norm as the constraint on perturbation, with  budget \(\epsilon = 16/255\). The learning rate is chosen as \(\alpha = 2/255\).
We use the Logit loss \cite{Logit} as our objective function in \Cref{eqn:objective}. By default,  we set the maximum number of optimization iterations to $1,000$ and the batch size to  $32$ for all baseline methods, to ensure sufficient optimization.  Furthermore, we notice that different attack methods may take different amounts of computations per iteration. We hence  fix the same number of inferences  for each optimization iteration to maintain fair comparisons across all attack methods.
\vspace{-1em}

\paragraph{Surrogate and Target Models} Our experiments choose various models commonly used in the literature \cite{CFM}. These include~\textbf{1)}  CNN-based models: VGG-16 \cite{VGG16}, ResNet-18 (RN18) \cite{ResNet}, ResNet-50 (RN50) \cite{ResNet}, DenseNet-121 (DN121) \cite{DenseNet}, Xception (Xcep) \cite{Xception}, MobileNet-v2 (MBv2) \cite{Mobilenetv2}, EfficientNet-B0 (EFB0) \cite{Efficientnet}, Inception ResNetv2 (IRv2) \cite{Inception-v4}, Inception-v3 (Incv3) \cite{Inception-v3}, and Inception-v4 (Incv4) \cite{Inception-v4}; and \textbf{2)} Transformer-based models: ViT \cite{ViT}, LeViT \cite{LeViT}, ConViT \cite{ConViT}, Twins \cite{Twins}, and Pooling-based Vision Transformer (PiT) \cite{PiT}. All the models are pre-trained on the ImageNet dataset \cite{Imagenet} . Additionally, we include CLIP \cite{CLIP}, trained on 400 million text-image pairs, to evaluate the transferability across different modalities. When using Transformer-based models, the input image is resized to 224$\times$224 to meet the model input requirement.

\vspace{-1em}
\input{tables/targeted_cnn_results.tex}

\paragraph{Baseline Attack Methods} We compare \name\ with various existing transfer-based methods, including DI \cite{DI}, RDI \cite{RDI}, SI \cite{SI}, Admix \cite{Admix},SIA \cite{SIA}, BSR \cite{BSR}, ODI \cite{ODI}, and CFM \cite{CFM}.
We also include two existing self-ensemble methods, namely, MUP  (whose implementation only handles CNN layers) \cite{MUP} and SE-ViT (which is specifically designed for vision Transformers)~\cite{SETR}.
These methods are primarily used in combination with TI-FGSM \cite{TI} and MI-FGSM \cite{MI} during the optimization process.
{It is worth noting that some of these methods, namely, SI, BSR, Admix, CFM, MUP and SE-ViT, are implemented together with RDI, which have been reported to obtain higher transfer ASRs \cite{CFM}.}   As previously mentioned, the attack methods are configured with an identical number of inferences per iteration, denoted as $S$. Specifically, we pick $S$ scaled copies in SI and in the inner loop  of Admix, and  $S$ transformed images for BSR . For other baselines, we perform $S$ forward-backward passes and use the average gradient on $x_{\mathrm{adv}}$ to update the adversarial example in each iteration. In the main experiment, we will set $S=5$; comparison of other choices is studied in \Cref{sec:differentS}.

\begin{figure*}[htbp]
    \centering
    
    \begin{minipage}{0.9\textwidth} 
        \centering
        \begin{subfigure}[b]{0.31\textwidth}
            \centering
            \captionsetup{labelformat=offset}
            \includegraphics[width=\linewidth]{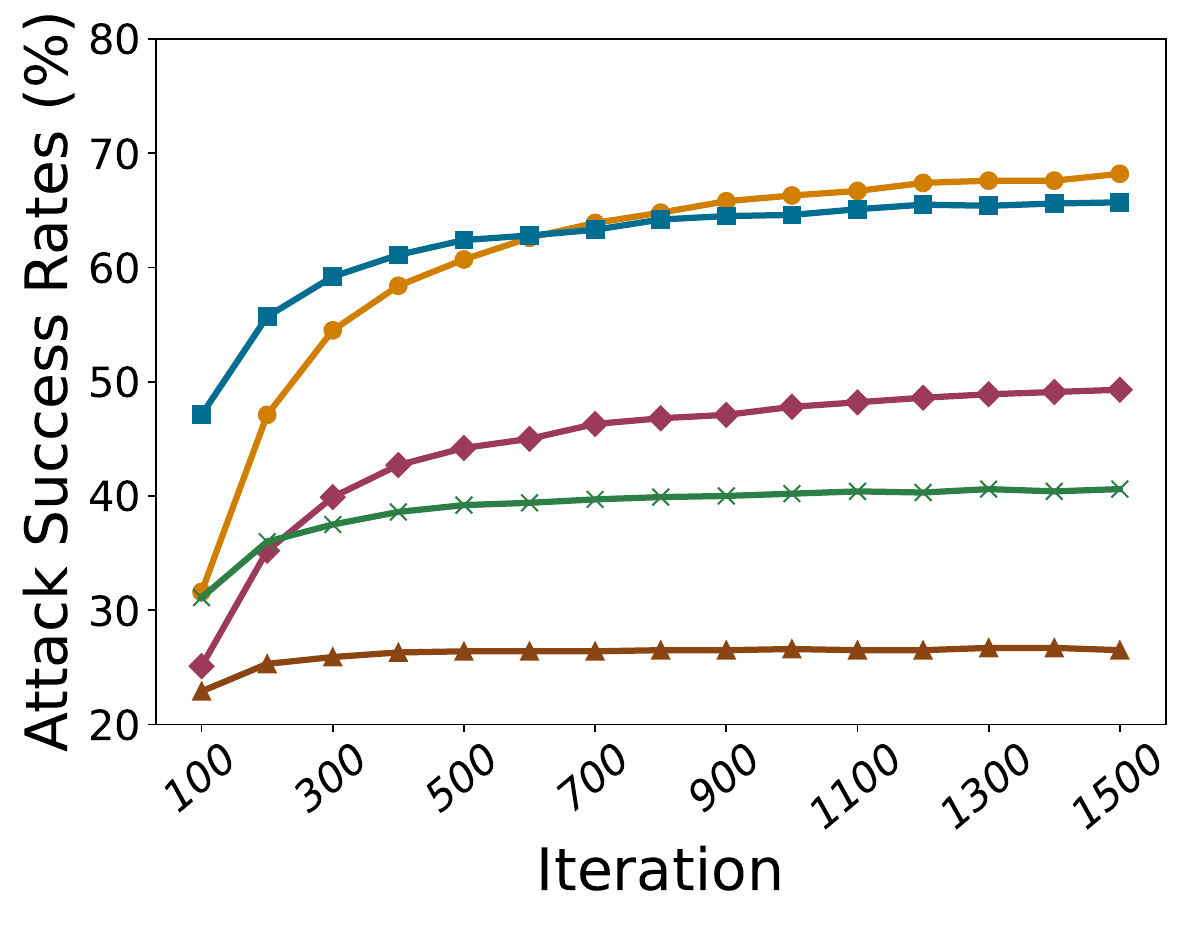}
            \caption{$S=1$}
        \end{subfigure} 
        \begin{subfigure}[b]{0.31\textwidth}
            \centering
            \captionsetup{labelformat=offset}
            \includegraphics[width=\linewidth]{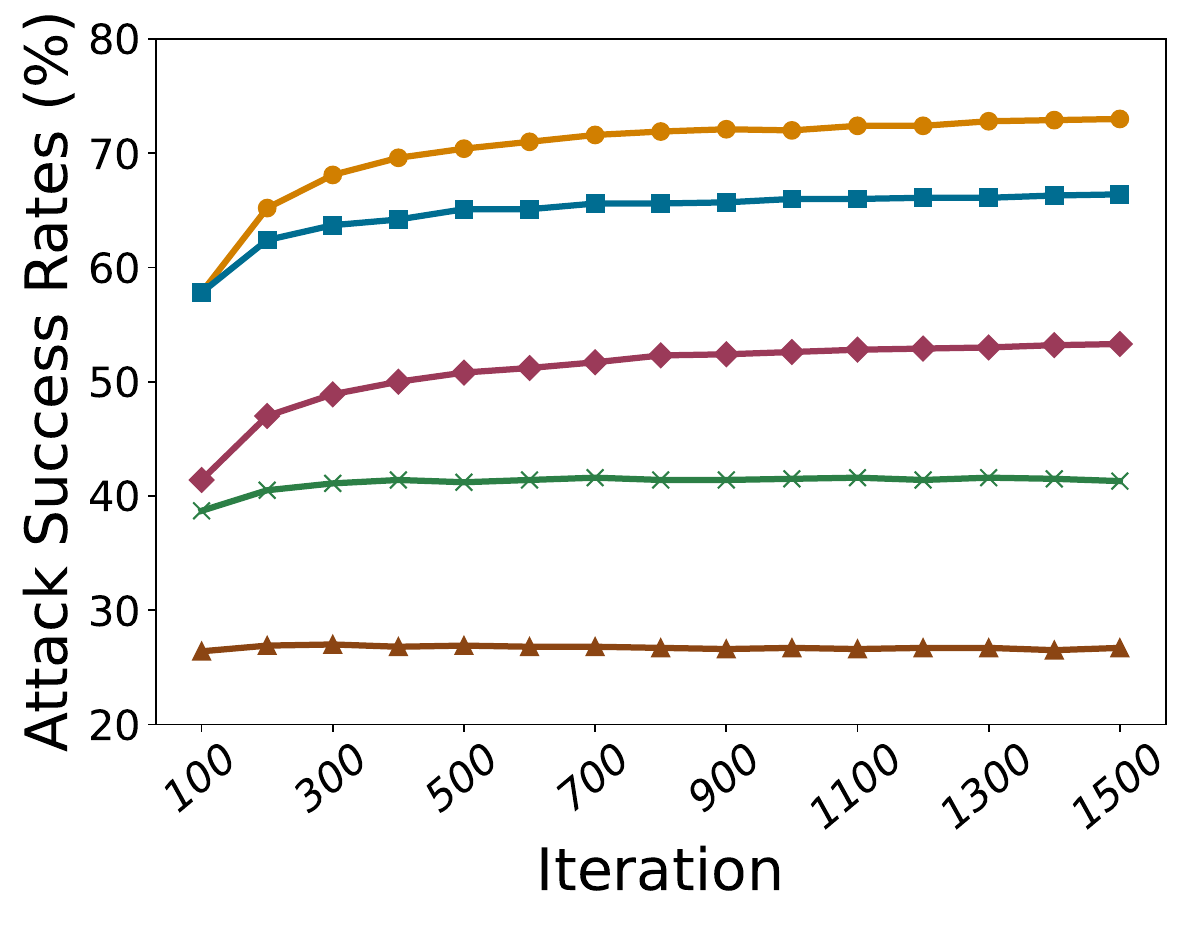}
            \caption{$S=5$}
        \end{subfigure} 
        \begin{subfigure}[b]{0.31\textwidth} 
            \centering
            \captionsetup{labelformat=offset}
            \includegraphics[width=\linewidth]{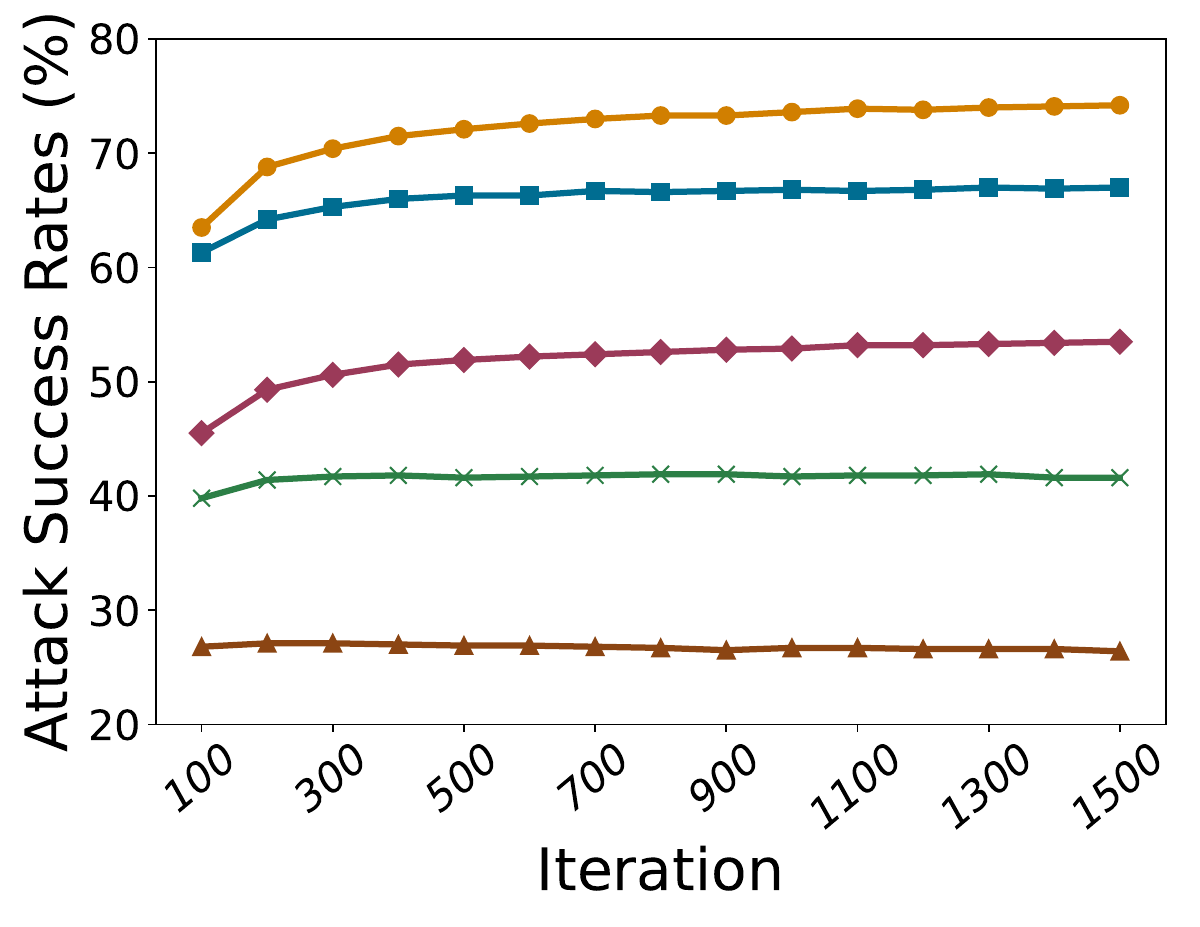}
            \caption{$S=10$}
        \end{subfigure}
    \end{minipage}
    \hspace{-0.025\textwidth} 
    \begin{minipage}{0.08\textwidth}
        \centering
        
        \includegraphics[width=\linewidth]{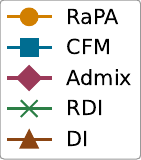}
    \end{minipage}
    \vspace{-0.5em}
    \caption{Average ASRs along optimization iterations. Here $S$ denotes the number of inferences per iteration.}
    \label{fig:iter/step}
\end{figure*}
\vspace{-1em}
\paragraph{\name\ Setting} While the DropConnect probabilities can be chosen differently for the weight and bias parameters in a linear layer (or the transformation parameters in the normalization layer) and also across different layers, we choose the same probability  for all selected parameters, that is, $p_w=p_b=p$, which greatly simplifies the implementation. 
Through our ablation study, we  find that \name\ performs well across a  range of probabilities.  For our experiments, we will select the following DropConnect probabilities: $0.05$ for ResNet-50, $0.02$ for Inception-v3, $0.04$ for DenseNet-121, $0.01$ for Vision Transformer, and $0.03$ for CLIP. By default, \name\ applies DropConnect to all linear and normalization layers in the surrogate model.

\subsection{Main Result}
\label{sec:mainresult}

We first study the performance of \name\ on the ImageNet-Compatible dataset. We employ ResNet-50, Inception-v3, DenseNet-121, and ViT as surrogate models, and evaluate the obtained adversarial examples on 16  target models.

\Cref{tab:targeted_vit_results} reports the experimental results when adversarial examples are generated using CNN-based models and  transferred to Transformer-based neural networks. This task is  considered more challenging in the context of transfer-based attacks \cite{mahmood2021robustness}, as the ASRs in this case are relatively low. Our method significantly improves the attack performance over existing methods: it increases the average ASR from 33.3\% to 45.0\% with ResNet-50 as surrogate model, and from 22.8\% to 40.3\% with DenseNet-121.

\Cref{tab:targeted_cnn_results} reports the  ASRs of various attack methods on ten CNN-based target models. \name\ achieves the best average ASR. Particularly, with Inception-v3 as surrogate model, the ASRs are increased by 14.6\% and 20.7\% for the challenging target models VGG16 and MBv2, respectively. When transferring from  Transformer-based model  ViT to CNN-based models, \name\ again attains the best average ASR  51.2\%. We also report the results  of  self-ensemble methods MUP \cite{MUP} and SE-ViT  \cite{SETR}. \name\ clearly outperforms these two methods by a large margin.

Additional experimental results can be found in \Cref{app:additionalresults}. We also visualize the heatmaps of some adversarial examples in \Cref{app:heatmap} for qualitative comparison.

\subsection{Ablation Study}
In this section, we conduct an ablation study to investigate the impacts of  1)  different  types of layers where DropConnect is applied, 2) different DropConnect probabilities and 3) more iterations and inferences.

\paragraph{Different Types of Layers}
We use ResNet-50 and ViT as surrogate models to analyze the impacts of layer types. For ResNet-50, we apply \name\ to different combinations of  Batch Normalization (BN) layer, Fully Connected (FC) layer, and Convolutional (Conv) layer. Similarly, we consider  Layer Normalization (LN)  and  FC layers (including  linear transformation layer in the attention layer) for ViT.

\Cref{tab:different_layer_type} presents the experimental results.    Notably, simply applying \name\ to all layers achieves equal or higher ASRs, compared with other baselines. The combination of BN (or LN) and FC layers performs the best, which validates our implementation in \Cref{sec:method}. For ResNet-50, applying DropConnect to Conv layers performs worse than  BN layers. We conjecture that  Conv layers have  sparser weights and may be less affected by over-reliance issue. Applying DropConnect only to FC layers yields particularly low ASRs, as ResNet contains only a single FC layer.
\input{tables/different_layer_type}

\vspace{-1em}
\paragraph{DropConnect Probability}

We investigate the impact of varying DropConnect probabilities $p$ using ResNet-50 as a surrogate model. \name\ is run with $p$ ranging from $0.01$ to $0.09$, and we report the average ASRs over 16 models in \cref{app:drop_prob}. The mean ASR is 66.3\% with a standard deviation of 5.9\%, peaking at 72.4\% when $p=0.05$. Notably, with $p \in [0.03, 0.07]$, \name\ consistently outperforms baselines by over 2\%, underscoring the stability of the proposed method across different choices.

\vspace{-1em}
\paragraph{More Iterations and Inferences}
\label{sec:differentS}
We study performance under different total iterations and numbers of inferences per iteration, denoted as $T$ and $S$, respectively. We use ResNet-50 as the surrogate model and evaluate how well adversarial examples transfer to the 16 target models.

\begin{figure}
    \centering
    \includegraphics[width=0.8\linewidth]{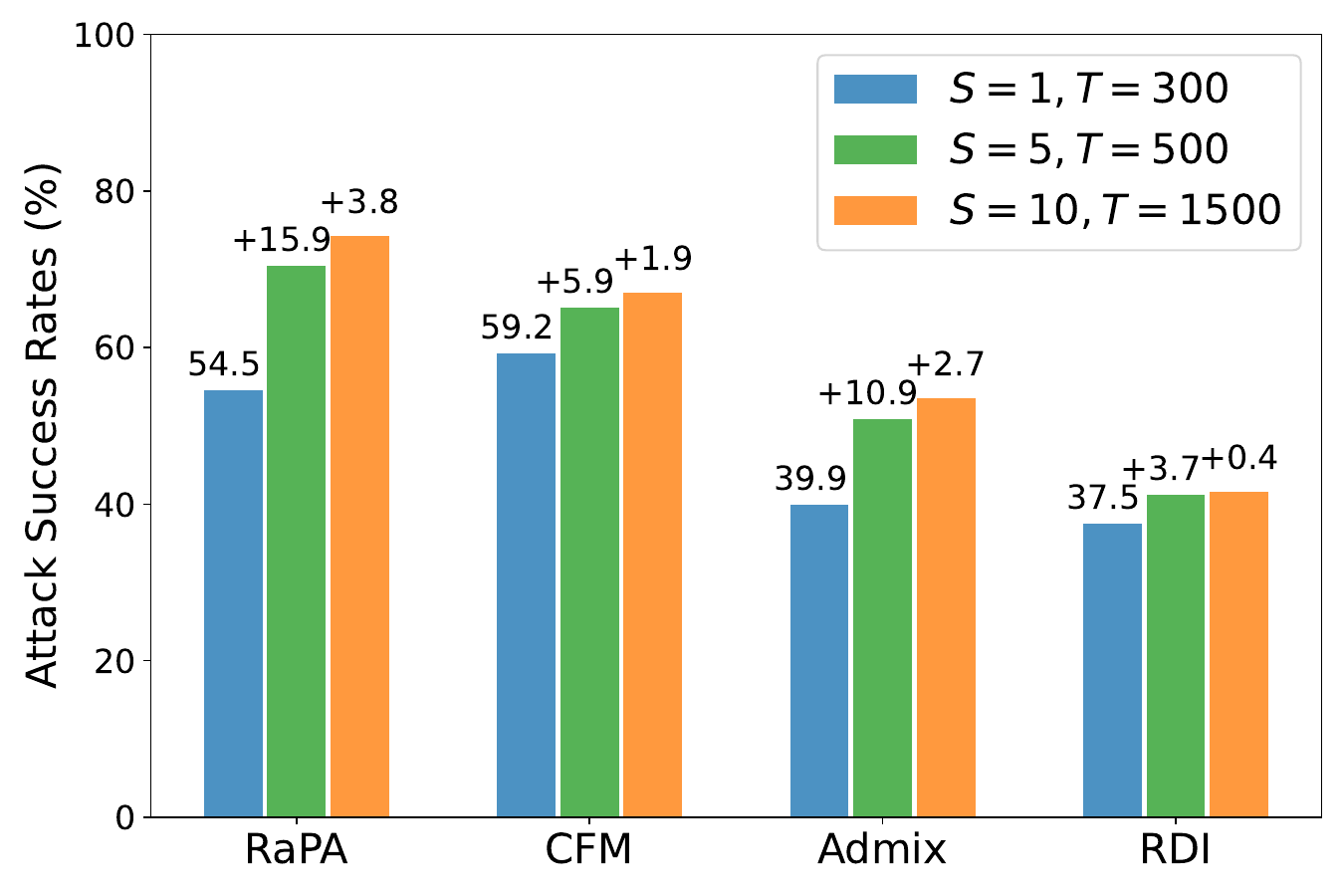}
    \vspace{-0.7em}
    \caption{Average ASRs with different  iterations  ($T$) and different numbers of inferences per iteration ($S$).}
    \label{fig:bar-chart}
    \vspace{-1.5em} 
\end{figure}

The results are reported in  \Cref{fig:iter/step}. Although existing methods may also benefit from additional optimization iterations, \name\ and Admix have the best gains when $T$ increases, while \name\ achieves a much higher ASR than Admix. With $S$ increasing, \name\ can outperform CFM even at an early stage of the optimization process. We also depict  \Cref{fig:bar-chart} to ease the comparison of the gains of different methods when both $T$ and $S$ increase. As we  observe, \name\ benefits the most from an additional compute budget.
\subsection{Attack Performance Against Defenses} 
We evaluate \name\ against several defenses: adversarially trained ResNet-50 (advRN) \cite{advResNet}, Ensemble-Adversarial-Inception-ResNet-v2 (ensIR) \cite{advIR}, High-level representation Guided Denoiser (HGD) \cite{HGD}, Bit Depth Reduction (Bit) \cite{JPEGandBit}, JPEG compression \cite{JPEGandBit}, R\&P \cite{R&P}, and Diffpure \cite{Diffpure}. We utilize ResNet-50 as the surrogate model. For Bit, JPEG, and R\&P, the target model is  ResNet-18 and for Diffpure, the target model is ResNet-50. As shown in \Cref{tab:targeted_defense},  \name\  outperforms all other baselines. Notably, against the strong defenses ensIR and HGD, \name\ exceeds the second-best ASRs by 29.4\% and 10.5\%, respectively.
\input{tables/targeted_defense}

\input{tables/compare_sasd_mba}

\subsection{ Training-enhanced Frameworks}

We compare RaPA with two training-dependent approaches, DSM\cite{DSM} and SASD-WS\cite{SASD-WS}, which involve additional optimization to enhance surrogate models for better adversarial transferability.
Specifically, DSM trains a surrogate model with dark knowledge extracted from a teacher model and enriched by mixing augmentation. SASD-WS improves transferability via sharpness-aware self-distillation and weight scaling, refining the loss landscape and model generalization.
 Table~\ref{tab:compare_sasd_mba} shows that under a fully training-free setting, RaPA already surpasses these training-dependent methods. Furthermore, when integrated with such training-based frameworks, RaPA continues to deliver consistent gains. For instance, combining with DSM increases the average ASR from 20.6\% to 58.3\%, highlighting its  compatibility  with existing training-enhanced frameworks.

\section{Concluding Remarks}

In this paper, we reveal the over-reliance issue in existing transfer-based attacks, 
where adversarial examples depend excessively on a small subset of model parameters. 
To alleviate this, we propose \name, which randomly prunes surrogate parameters during optimization. 
We show that the expected effect of random pruning equals adding an importance-equalization regularizer, thereby reducing parameter over-reliance and improving transferability. 
Extensive experiments on CNN and Transformer architectures confirm the effectiveness and stability of \name.

%% file: tables/targeted_vit_results.tex
\begin{table*}[!ht]
\centering
\renewcommand{\arraystretch}{1.0}
\resizebox{0.92\textwidth}{!}{%
\begin{tabular}{l|ccccccc|ccccccc}
\toprule[0.15em]
 \multirow{1}{*}{} & \multicolumn{7}{c|}{\textbf{Source: RN50}} & \multicolumn{7}{c}{\textbf{Source: DN121}} \\ 
 \cmidrule(l){2-8} \cmidrule(l){9-15}
Attack & ViT & LeViT & ConViT & Twins & PiT & CLIP & Avg. & ViT & LeViT & ConViT & Twins & PiT & CLIP & Avg. \\ \midrule
DI & 0.4 & 6.7 & 0.6 & 3.8 & 1.8 & 0.5 & 2.3   & 0.3 & 4.0 & 1.2 & 2.0 & 2.1 & 0.3 & 1.7 \\
RDI & 2.8 & 24.0 & 4.4 & 12.6 & 10.1 & 1.2 & 9.2   & 1.0 & 12.0 & 2.1 & 6.9 & 8.2 & 1.4 & 5.3 \\
SI & 8.0 & 42.9 & 7.7 & 25.1 & 23.3 & 3.7 & 18.4   & 3.9 & 21.5 & 4.4 & 10.2 & 14.3 & 2.1 & 9.4 \\
SIA & 3.1 & 28.4 & 3.9 & 16.7 & 13.5 & 1.9 & 11.2 & 2.4 & 24.2 & 2.2 & 12.9 & 11.1 & 1.5 & 9.0 \\
BSR & 6.8 & 42.4 & 7.7 & 25.3 & 21.9 & 2.3 & 17.7  & 2.7 & 21.6 & 2.7 & 10.6 & 11.9 & 1.2 & 8.5 \\
DWP & 3.7 & 32.0 & 4.0 & 17.6 & 13.3 & 1.8 & 12.1 & 2.7 & 21.1 & 2.9 & 14.3 & 12.0 & 2.4 & 9.2 \\
Admix & 7.2 & 43.4 & 5.9 & 22.7 & 19.4 & 4.2 & 17.1  & 4.2 & 33.8 & 4.0 & 18.0 & 19.1 & 3.9 & 13.8 \\
ODI & 15.5 & 49.3 & 11.8 & 31.6 & 35.1 & 5.6 & 24.8  & 8.5 & 36.3 & 8.4 & 19.8 & 26.7 & 4.9 & 17.4 \\
MUP & 7.0 & 48.3 & 8.2 & 30.7 & 26.5 & 3.6 & 20.7 & 5.2 & 38.9 & 5.0 & 23.2 &20.1 &4.1 & 16.1 \\
CFM & 17.3 & 65.8 & 14.6 & \underline{47.5} & 39.9 & 7.9 & 32.2 & 10.4 & 49.0 & 8.3 & 30.1 & \underline{30.4} & \underline{5.5} & 22.3 \\
FTM & \underline{18.0} & \underline{67.6} & \underline{16.5} & 47.1 & \underline{41.5} & \underline{9.3} & \underline{33.3} & \underline{10.7}&\underline{50.4}&\underline{9.4}&\underline{31.1}&\underline{30.4}&4.8&\underline{22.8}
\\
\rowcolor{Gray} \name & \textbf{33.8}\raisebox{-0.7ex}[\height][0pt]{±1.0} & \textbf{75.4}\raisebox{-0.7ex}[\height][0pt]{±1.8} & \textbf{27.6}\raisebox{-0.7ex}[\height][0pt]{±0.2} & \textbf{59.5}\raisebox{-0.7ex}[\height][0pt]{±0.4} & \textbf{57.3}\raisebox{-0.7ex}[\height][0pt]{±0.7} & \textbf{15.6}\raisebox{-0.7ex}[\height][0pt]{±0.1} & \textbf{45.0}  & \textbf{27.8}\raisebox{-0.7ex}[\height][0pt]{±0.1} & \textbf{69.4}\raisebox{-0.7ex}[\height][0pt]{±0.6} & \textbf{23.5}\raisebox{-0.7ex}[\height][0pt]{±0.2} & \textbf{53.1}\raisebox{-0.7ex}[\height][0pt]{±0.6} & \textbf{54.0}\raisebox{-0.7ex}[\height][0pt]{±0.1} & \textbf{14.1}\raisebox{-0.7ex}[\height][0pt]{±0.0} & \textbf{40.3} \\
\bottomrule[0.15em]
\end{tabular}%
}
\vspace{-0.3em}
\caption{ASRs (\%) against five Transformer-based target models on the ImageNet-Compatible dataset. All the attack methods are combined with MI-TI. The best results are shown in bold and the second best results are underlined.}
\label{tab:targeted_vit_results}
\vspace{-0.7em}
\end{table*}

%% file: tables/targeted_cnn_results.tex
\begin{table*}[!ht]
\centering
\normalsize
\renewcommand{\arraystretch}{1.0}
\resizebox{0.85\textwidth}{!}{%
\begin{tabular}{lccccccccccc}
\toprule[0.15em]
\textbf{\begin{tabular}[c]{@{}l@{}}Source : Incv3\end{tabular}} & \multicolumn{10}{c}{Target model} &  \\ \cmidrule(l){2-12}
Attack & RN18 & RN50 & VGG16 & Incv3 & EFB0 & DN121 & MBv2 & IRv2 & Incv4 & Xcep & Avg. \\ \midrule
DI & 2.2 & 3.9 & 3.4 & \underline{99.1} & 3.6 & 5.0 & 1.2 & 7.7 & 8.9 & 7.0 & 14.2 \\
RDI & 5.8 & 5.5 & 3.9 & 99.0 & 8.0 & 8.5 & 3.8 & 18.6 & 18.8 & 11.1 & 18.3 \\
SI & 6.7 & 6.7 & 4.3 & 98.8 & 9.7 & 9.7 & 4.4 & 23.3 & 22.1 & 13.6 & 19.9 \\
MUP & 13.9 & 13.6 & 9.6	& 98.4	&17.7&22.4&8.1&42.2&42.2&26.4&29.5 \\
BSR & 15.8 & 13.9 & 11.9 & 98.7 & 20.5 & 24.3 & 9.6 & 45.7 & 45.5 & 30.3 & 31.6 \\
DWP & 17.5 & 17.2 & 13.5 & 99 & 19.2 & 29.8 & 9.7 & 54.4 & 52.9 & 37.5 & 35.1 \\
Admix & 18.5 & 16.7 & 13.8 & 98.1 & 23.8 & 27.5 & 15.9 & 46.3 & 47.1 & 38.9 & 34.7 \\
SIA & 17.4 & 21.5 & 16.7 & 98.8 & 27.2 & 32.2 & 13.1 & 56.1 & 59.2 & 42.9 & 38.5 \\
ODI & 14.4 & 22.3 & 22.0 & \textbf{99.4} & 26.0 & 39.5 & 13.9 & 51.8 & 60.7 & 44.7 & 39.5 \\
CFM & \underline{37.4} & \underline{37.9} & \underline{27.3} & 97.9 & \underline{46.1} & \underline{53.0} & \underline{27.8} & \underline{76.9} & \underline{76.1} & \underline{68.2} & \underline{54.9} \\
\rowcolor{Gray}
\name & \textbf{51.3}\raisebox{-0.7ex}[\height][0pt]{±0.6} & \textbf{53.5}\raisebox{-0.7ex}[\height][0pt]{±1.0} & \textbf{41.9}\raisebox{-0.7ex}[\height][0pt]{±0.6} & 97.4\raisebox{-0.7ex}[\height][0pt]{±0.0} & \textbf{60.8}\raisebox{-0.7ex}[\height][0pt]{±0.3} & \textbf{68.4}\raisebox{-0.7ex}[\height][0pt]{±1.4} & \textbf{48.5}\raisebox{-0.7ex}[\height][0pt]{±0.0} & \textbf{86.7}\raisebox{-0.7ex}[\height][0pt]{±0.3} & \textbf{87.5}\raisebox{-0.7ex}[\height][0pt]{±0.4} & \textbf{84.0}\raisebox{-0.7ex}[\height][0pt]{±0.0} & \textbf{68.0} \\

\midrule
\textbf{\begin{tabular}[c]{@{}l@{}}Source : ViT\end{tabular}} & \multicolumn{10}{c}{Target model} &  \\ \cmidrule(l){2-12}
Attack & RN18 & RN50 & VGG16 & Incv3 & EFB0 & DN121 & MBv2 & IRv2 & Incv4 & Xcep & Avg. \\ \midrule
DI & 0.5 & 1.0 & 0.8 & 2.1 & 2.1 & 1.1 & 1.3 & 1.8 & 1.6 & 1.4 & 1.4 \\
RDI & 1.7 & 2.4 & 1.3 & 4.1 & 6.3 & 4.1 & 2.7 & 5.8 & 4.9 & 4.6 & 3.8 \\
SI & 2.9 & 3.9 & 1.1 & 8.0 & 9.2 & 5.9 & 2.7 & 7.9 & 6.2 & 6.8 & 5.5 \\
BSR & 5.0 & 8.2 & 3.9 & 11.2 & 15.0 & 12.8 & 5.3 & 15.5 & 13.4 & 11.2 & 10.2 \\
DWP & 13.5 & 11.7 & 7.1 & 15.8 & 22.2 & 16.7 & 10.2 & 17.4 & 16.3 & 16.3 & 14.7 \\
SE-ViT  & 8.9  & 12.0 & 6.9  & 21   & 25.3 & 20.5 & 8.3  & 23.8 & 22.5 & 20.3 & 17.0 \\
Admix & 16.4 & 20.4 & 13.2 & 31.8 & 38.3 & 31.5 & 17.8 & 35.3 & 31.9 & 29.8 & 26.6 \\
SIA & 3.6 & 4.8 & 3.0 & 8.1 & 12.2 & 8.4 & 3.5 & 8.9 & 10.1 & 8.3 & 7.1 \\
ODI & 12.4 & 20.0 & 10.6 & 28.3 & 28.9 & 30.9 & 10.4 & 35.5 & 34.4 & 27.9 & 23.9 \\
CFM & \underline{26.1} & \underline{33.4} & \underline{18.0} & \underline{45.2} & \underline{56.8} & \underline{47.3} & \underline{23.2} & \underline{54.5} & \underline{49.9} & \underline{46.2} & \underline{40.1} \\
\rowcolor{Gray}
\name & \textbf{37.2}\raisebox{-0.7ex}[\height][0pt]{±0.7} & \textbf{42.9}\raisebox{-0.7ex}[\height][0pt]{±0.7} & \textbf{28.6}\raisebox{-0.7ex}[\height][0pt]{±2.0}
& \textbf{57.5}\raisebox{-0.7ex}[\height][0pt]{±1.4} & \textbf{68.2}\raisebox{-0.7ex}[\height][0pt]{±0.4} & \textbf{56.8}\raisebox{-0.7ex}[\height][0pt]{±0.2} & \textbf{36.4}\raisebox{-0.7ex}[\height][0pt]{±0.7} & \textbf{62.9}\raisebox{-0.7ex}[\height][0pt]{±0.4} & \textbf{62.6}\raisebox{-0.7ex}[\height][0pt]{±0.1} & \textbf{58.3}\raisebox{-0.7ex}[\height][0pt]{±1.0} & \textbf{51.2} \\
\bottomrule[0.15em]
\end{tabular}%
}
\vspace{-0.3em}
\caption{ASRs (\%) against ten target models on the ImageNet-Compatible dataset. All the attack methods are combined with MI-TI. The best results are shown in bold and the second best results are underlined.}
\label{tab:targeted_cnn_results}
\end{table*}

%% file: tables/different_layer_type.tex
\begin{table}[h]
\centering
\resizebox{0.39\textwidth}{!}{%
    \begin{tabular}{lcccc}
    \toprule[0.15em]
    \textbf{Source Model} & \textbf{BN} & \textbf{FC} & \textbf{Conv} & \textbf{ASRs (\%)} \\ \midrule
    \multirow{7}{*}{RN50}  
                           & \checkmark   &               &             & \underline{72.1}   \\ 
                           &              &  \checkmark   &             & 41.1               \\
                           &              &               & \checkmark  & 65.1               \\ 
                           & \checkmark   & \checkmark    &             & \textbf{72.4}       \\ 
                           & \checkmark   &               & \checkmark  & 69.1               \\ 
                           &              & \checkmark    & \checkmark  & 64.7               \\ 
                           & \checkmark   & \checkmark    & \checkmark  & 69.2               \\ 
    
                           \midrule
    \textbf{Source Model}  & \textbf{LN}  & \textbf{FC}   &   -        & \textbf{ASRs (\%)} \\ \midrule
    \multirow{3}{*}{ViT}    & \checkmark   &               &             & 58.6               \\
                           &              & \checkmark    &             & \underline{63.9}               \\ 
                           & \checkmark   & \checkmark    &             & \textbf{65.2}               \\ 
    \bottomrule[0.15em]
    \end{tabular}
}
\vspace{-0.3em}
\caption{Applying DropConnect to different types of layers. The reported ASRs(\%) are averages  over 16  target models.}
\label{tab:different_layer_type}
\vspace{-0.7em}
\end{table}

%% file: tables/targeted_defense.tex
\begin{table}[!ht]
\centering
\renewcommand{\arraystretch}{1.0}
\resizebox{0.46\textwidth}{!}{%
\begin{tabular}{lcccccccc}
\toprule[0.15em]
\textbf{RN50} & \multicolumn{7}{c}{Defense methods} \\ \cmidrule(l){2-8}
Method     & advRN & ensIR & JPEG & Bit  & R\&P & HGD & Diffpure \\ 
\midrule
DI         & 10.6     & 0.0            & 26.7 & 50.5 & 41.1 & 0.1 & 0.0 \\
RDI        & 39.6     & 0.8          & 58.8 & 75.8 & 71.3 & 0.8 & 0.0\\ 
SI     & 61.8     & 9.4          & 75.1 & 80.7 & 78.8 & 2.1 & 0.4\\
Admix  & 68.9     & 5.7          & 76.9 & 81.4 & 78.6 & 2.4 & \underline{0.7} \\ 
BSR    & 60.4     & 3.0          & 76.8 & 85.2 & 83.4 & 2.6 & 0.1 \\ 
ODI        & 58.6     & 5.1          & 71.7 & 73.8 & 76.1 & 0.3 & 0.3\\
CFM    & \underline{84.1}     & \underline{13.8}         & \underline{88.2} & \underline{91.5} & \underline{89.3} & \underline{15.2} & 0.5\\
\rowcolor{Gray}
\name  & \textbf{88.2}     & \textbf{43.2}         & \textbf{91.2} & \textbf{92.2} & \textbf{92.7} & \textbf{25.7} & \textbf{4.0}\\ 
\bottomrule[0.15em]
\end{tabular}%
}
\vspace{-0.3em}
\caption{ASRs (\%) against six defense methods. }
\label{tab:targeted_defense}
\end{table}

%% file: tables/compare_sasd_mba.tex
\begin{table}[!ht]
\centering
\renewcommand{\arraystretch}{1.0}
\resizebox{0.46\textwidth}{!}{%
\begin{tabular}{lccccccc}
\toprule[0.15em]
Attack & ViT & LeViT & ConViT & Twins & PiT & CLIP & Avg. \\
\midrule
\rowcolor{Gray}
RaPA & 33.8 & 75.4 & 27.6  & 59.5 & 57.3 & 15.6 & 45.0  \\
DSM & 8.1 & 49.5 & 7.8 & 31.6  & 23.3 & 3.1 & 20.6 \\
\rowcolor{Gray}
DSM-RaPA & 50.8 & 84.2 & 42.2 & 74.6 &72.6 &25.1 & 58.3 \\
SASD-WS & 30.1 &74.1 &25.6 &52.1 &52.9 &18.1 &42.2 \\
\rowcolor{Gray}
 SASD-RaPA & 42.9 & 79.5 &35.3 &63.0 &63.1 &25.9 &51.6 \\
\bottomrule[0.15em]
\end{tabular}%
}
\vspace{-0.3em}
\caption{ASRs (\%) of adversarial examples generated by ResNet50 against Transformer-based targets, comparing RaPA with training-based methods (DSM and SASD-WS) and their combinations. Note that when RaPA is combined with SASD-WS, we did not apply Weight Scaling.}
\label{tab:compare_sasd_mba}
\vspace{-0.7em}
\end{table}

%% file: sec/acknowledge.tex
\section{Acknowledgement}
This work was supported by the Strategic Priority Research Program of the Chinese Academy of Sciences (Grant No. XDB0680101); the National Key Research and Development Program of China (Grant No. 2023YFA1011602); the CAS Project for Young Scientists in Basic Research (Grant No. YSBR-034); the Xiaomi Young Talents Program; and the Innovation Project of Institute of Computing Technology, Chinese Academy of Sciences (Grant No.~E561130).

%% file: sec/X_supplementary.tex
\appendix
\setcounter{secnumdepth}{2}     
\renewcommand\thesubsection{\thesection.\arabic{subsection}}
\renewcommand\labelenumi{\arabic{enumi}.}                  

\newpage
\onecolumn

\begin{center}
    {\LARGE \bf Supplementary Material}
\end{center}

\section{More Related Work}
\label{app:related_work}

One of the most fundamental attack methods is Fast Gradient Sign Method (FGSM) \cite{FGSM}, which uses the direction of gradient to craft adversarial examples. Iterative-FGSM (I-FGSM) \cite{I-FGSM} extends FGSM into an iterative framework to enhance the attack performance. However, while the obtained adversarial examples achieve high success rates for white-box attacks, they do not perform well when transferred to other black-box models. To improve  transferability, many  methods have been proposed, mainly from the following  perspectives: input transformation, gradient stabilization, loss function refinement, and model ensemble.

The first class applies   transformations to the input images, thereby diversifying the input patterns \cite{DI,RDI,TI,ODI,Admix,BSR,CFM}.
Diverse Inputs (DI) \cite{DI} is one of the representative methods. At each iteration, it applies
random and differentiable transformations (e.g., random resizing) to the input image  and then maximizes the loss function w.r.t.~the
  transformed inputs.     Resized Diverse Inputs (RDI) \cite{RDI} extends DI by resizing the transformed image back to its original size.
The Translation-Invariant (TI) attack method \cite{TI} optimizes perturbations across a set of translated images, making the  adversarial example less sensitive to the surrogate model. 
Block Shuffle and Rotate (BSR) \cite{BSR} divides the input image into blocks, followed by shuffling or rotation, and 
{Object-based Diversity Inputs (ODI)  \cite{ODI} generates an adversarial example  on a 3D object and leads the rendered image to being classified as the target class.} 
Admix  \cite{Admix} mixes the input image with random samples from other classes. Clean Feature Mixup (CFM) \cite{CFM} extends Admix to high-level feature space and introduces two types of competing noises to guide adversarial perturbations. Building upon it, Feature Tuning Mixup (FTM) \cite{FTM} further introduces learnable and attack-specific feature perturbations that combine random and optimized noises in the feature space, achieving SoTA performance in transfer-based targeted attacks.

The second class of methods stabilize the gradient updates to enhance transferability during adversarial example generation.
Momentum Iterative FGSM (MI-FGSM) \cite{MI} incorporates a momentum term into I-FGSM, helping avoid local optima.
The Variance-Tuned (VT)  method \cite{VT} further improves stability by not only accumulating the gradients at current iteration but also incorporating the variance of gradients from previous iterations to adjust the current update. 
Scale-Invariant (SI) optimization \cite{SI}  leverages the scale-invariance property of deep learning models by applying perturbations across multiple scaled copies of the input image, in order to reduce overfitting to the white-box model.

Another direction is to devise tailored loss functions, particularly for targeted attacks. The Po+Trip method \cite{Po+trip} introduces  Poincar\'{e} distance as a similarity measure and dynamically adjusts gradient magnitudes to mitigate the ``noise curing" issue. 
The Logit method \cite{Logit}, which directly maximizes the logit output of the target class, alleviates the gradient vanishing problem and has demonstrated significant improvement w.r.t.~transferability of targeted attack.

Additionally, surrogate model ensemble is also useful to enhancing  transferability, where one utilizes the average of losses, predictions, or logits of multiple  models to craft adversarial examples \cite{MI,Liu2016DelvingIT}. A recent work \cite{Chen2023RethinkingME} proposes the common weakness attack method composed of sharpness aware minimization  and cosine similarity encourager, beyond the averaging. As shown in \cite{T-sea,liu2024scaling},  the transferability generally improves with an increasing number of surrogate classifiers. However, the number of surrogate models in practice is usually small and picking proper surrogate models for the same task is also not easy.

Another line of work focuses on improving the \emph{surrogate model itself} through additional training to enhance the adversarial transferability.
Dark Surrogate Model (DSM) \cite{DSM} trains a surrogate model using \emph{dark knowledge} distilled from a teacher model and further augments the training data with \emph{mixing augmentations} such as CutMix and Mixup, thereby enriching soft supervision and improving transferability.
SASD-WS \cite{SASD-WS} introduces \emph{Sharpness-Aware Self-Distillation (SASD)} to flatten the loss landscape and combines it with \emph{Weight Scaling (WS)} to approximate model ensembling.
While these approaches require retraining surrogate models, our RaPA works in a \emph{training-free} manner, yet achieves comparable or even superior transferability, and can be seamlessly integrated with such training-based frameworks.

Closely related to the present work is self-ensemble \cite{T-sea,MUP,Ghost,SETR}, which creates
multiple models from only one surrogate model. Transfer-based self-ensemble (T-sea) \cite{T-sea} locally apply various transformations onto the input image to improve such diversity while preserving the structure of imagetargets the task of object detection, and the self-ensemble method in  \cite{SETR} specifically considers vision Transformer as surrogate model and is denoted as SE-ViT in this paper. Ghost Network \cite{Ghost} perturbs surrogate model to create a set of new models and then samples one model from the set at each iteration. Masking Unimportant Parameters (MUP) \cite{MUP} drops out unimportant parameters according to a predefined Taylor expansion-based metric.
\section{Estimation and Approximation}
\subsection{Parameter Importance Estimation and Pruning by Importance}
\label{sec:prune_method}
Computing the second-order derivative defined in Eq.~\eqref{eqn:importance} is generally costly. A more compact approximation is obtained via first-order expansion \cite{molchanov2019importance}, which reduces to

\begin{align}
      \mathcal{I}(\theta_i) = \frac{\partial^2 \mathcal{L}\left(f(x_\mathrm{adv})\right)}{\partial \theta_i^2} \theta_i^2
      \approx \left(\frac{\partial \mathcal{L}\left(f(x_\mathrm{adv})\right)}{\partial \theta_i}\theta_i\right)^2.
      \label{eqn:approx_importance}
\end{align}
In our pilot study, we adopt the method in DepGraph \cite{fang2023depgraph} to perform the pruning, which explicitly models inter-layer dependencies and comprehensively groups coupled parameters for pruning.

\subsection{Derivation of the Masked Loss Approximation in Eq.~\eqref{eqn:taylor_mask}}
Let ${\theta}$ represent the entire set of model parameters. We define a random binary mask $\mathcal{M} \in \{0,1\}^{|\theta|}$, where each entry is independently sampled from a Bernoulli distribution: $\mathcal{M}_i \sim \text{Bernoulli}(1 - p)$.  A second-order Taylor expansion gives
\begin{align}
\mathcal{L}\bigl(f(x_{\mathrm{adv}}; \mathcal{M}\odot\theta)\bigr) \nonumber 
\approx &~\mathcal{L}\bigl(f(x_{\mathrm{adv}}; \theta)\bigr) 
+ \sum_i g_i \Delta_i
+ \tfrac{1}{2} \sum_{i,j} H_{ij} \Delta_i \Delta_j,
\end{align} where $\Delta = (\mathcal{M}-\mathbf{1}) \odot \theta$, $g_i=\partial \mathcal{L}\bigl(f(x_{\mathrm{adv}};\theta)\bigr)/\partial \theta_i$ and $H_{ij}=\partial^2 \mathcal{L}\bigl(f(x_{\mathrm{adv}};\theta)\bigr)/(\partial \theta_i \partial \theta_j)$ evaluated at $\theta$. Since  $p$ is taken to be small and  close to zero, we have $\mathbb{E}[\Delta_i]\approx 0$. And the mask entries are independent, so we further have $\mathbb{E}[\Delta_i\Delta_j]\approx 0$ for $i\neq j$ while $\mathbb{E}[\Delta_i^2]=p(1-p)\theta_i^2$. Taking expectations yields
\begin{align}
\mathbb{E}_{\mathcal{M}}\!\left[\mathcal{L}\bigl(f(x_{\mathrm{adv}}; \mathcal{M}\odot\theta)\bigr)\right]
\approx \mathcal{L}\bigl(f(x_{\mathrm{adv}}; \theta)\bigr) + 
\frac{p (1-p) }{2} \sum_i H_{ii}\,\theta_i^2,
\label{eqn:taylor_mask_appendix}
\end{align}
which is the expression  in Eq.~\eqref{eqn:taylor_mask}.
\subsection{Detailed Gini Coefficients Computation Procedure}
\label{app:detailed_gini}

We compute the Gini coefficients based on the parameter importance values $\mathcal{I}(\theta_i)$ defined in Eq.~\eqref{eqn:importance}. 
For each layer $l$ in the surrogate model, we first collect all parameter importance values $\{\mathcal{I}^l(\theta_i)\}_{i=1}^n$. 
To better highlight the disparity among parameters, we further apply an exponential scaling to these importance values before computing the Gini coefficient:
\begin{equation}
\tilde{\mathcal{I}}^l(\theta_i) = \exp\left( \mathcal{I}^l(\theta_i) \right),
\label{eqn:importance_exp_scaled}
\end{equation}
where $\tilde{\mathcal{I}}^l(\theta_i)$ denotes the scaled importance. 
This transformation magnifies the relative differences among parameters, enabling a clearer assessment of importance inequality.

Given the scaled importance values $\{\tilde{\mathcal{I}}^l(\theta_1), \tilde{\mathcal{I}}^l(\theta_2), \dots, \tilde{\mathcal{I}}^l(\theta_n)\}$ in layer $l$, we compute the Gini coefficient as:
\begin{equation}
\text{Gini}^l = \frac{2 \sum_{i=1}^{n} i \tilde{v}_{(i)}}{n \sum_{i=1}^{n} \tilde{v}_{(i)}} - \frac{n + 1}{n},
\label{eqn:gini_formula}
\end{equation}
where $\tilde{v}_{(i)}$ represents the $i$-th smallest value after sorting in ascending order.  
A smaller Gini coefficient indicates a more balanced distribution of parameter importance within the layer.

Finally, we aggregate the Gini coefficients across layers to obtain both the type-wise and overall average results:
\begin{equation}
\text{Gini}_{\text{avg}} = \frac{1}{|\mathbb{L}|} \sum_{l \in \mathbb{L}} \text{Gini}^l,
\label{eqn:gini_avg}
\end{equation}
where $\mathbb{L}$ denotes the set of all layers containing learnable parameters.  
This process allows us to quantitatively evaluate the degree of importance imbalance and verify that the proposed random parameter pruning strategy effectively reduces the model’s over-reliance on a small subset of dominant parameters.

\section{Visualization of Adversarial Examples}
\label{app:heatmap}

\begin{figure*}[ht!]
    \centering
    
    \includegraphics[width=1\linewidth]{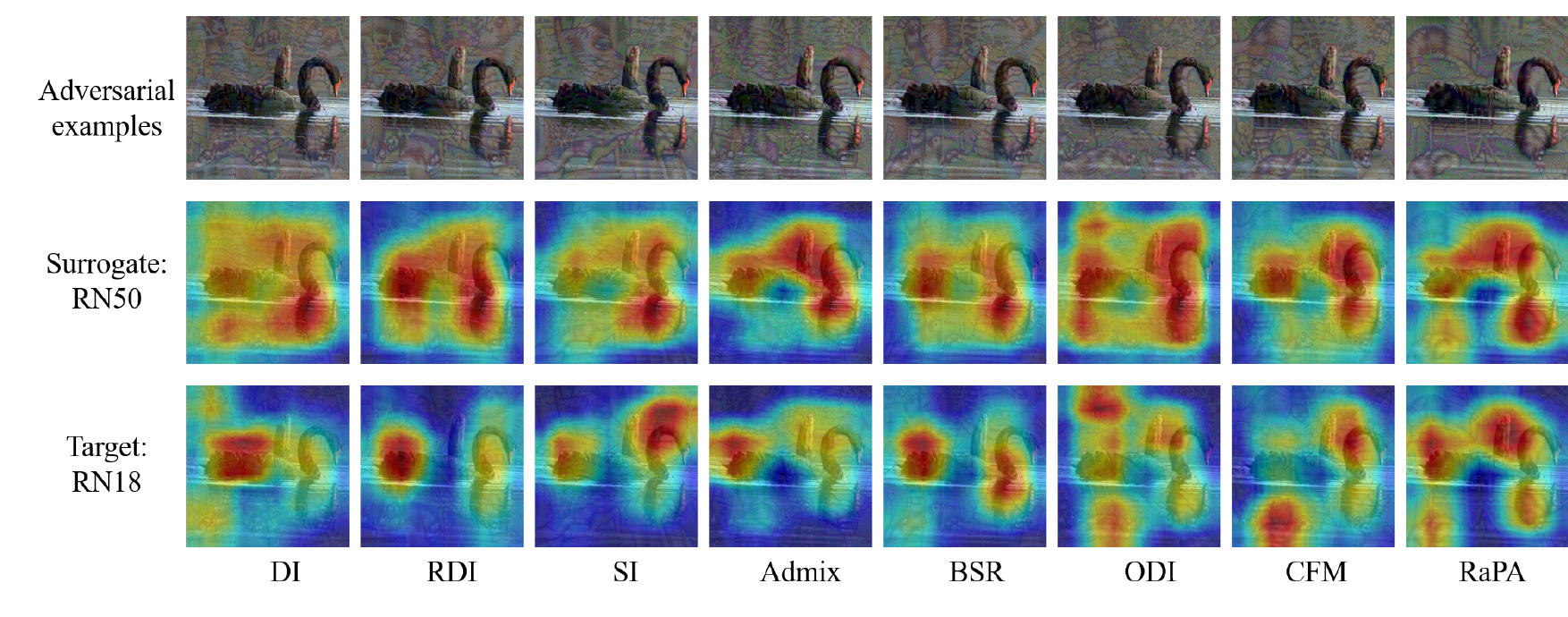}
    
    \caption{Attention heatmap of adversarial example. The true label is `black swan' and the target label is `weasel'. The intensity of red indicates the level of importance assigned to each area, influencing the classifier prediction towards the target label.}
    \label{fig:black swan_weasel}
    
\end{figure*}

\begin{figure*}[ht!]
    \centering
    
    \includegraphics[width=1\linewidth]{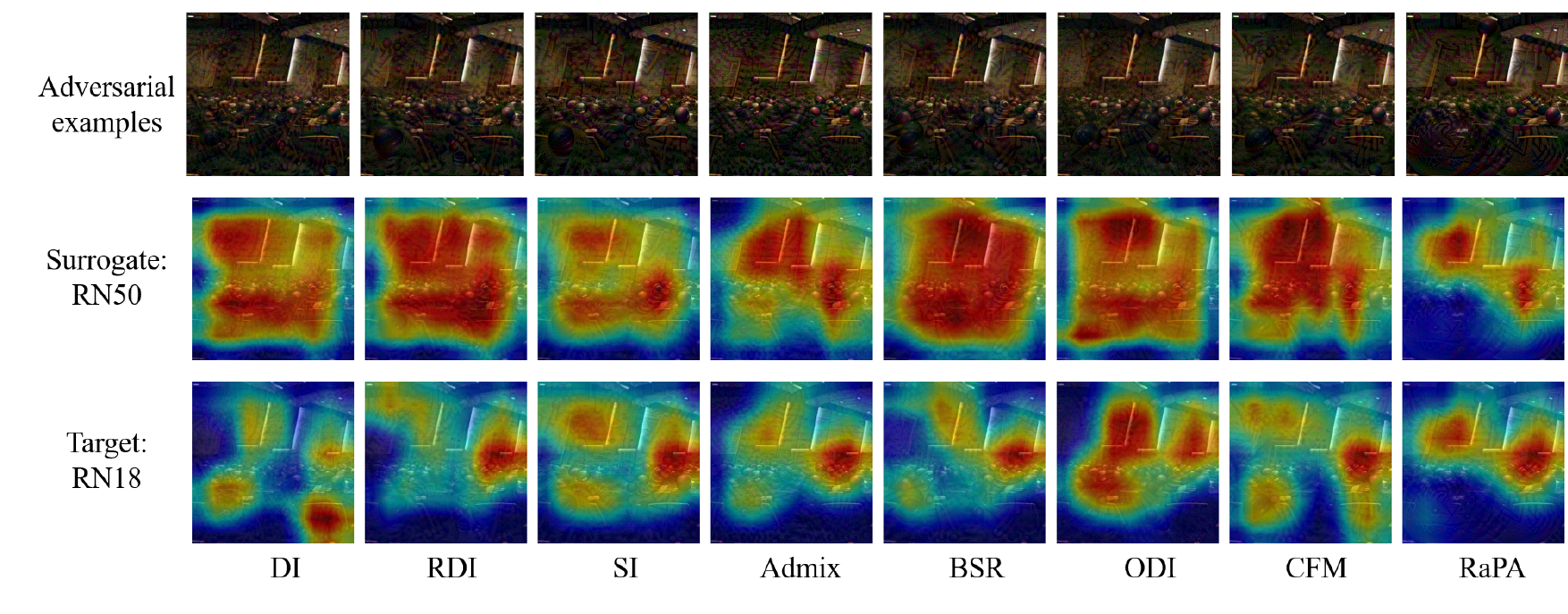}
    
    \caption{
    Attention heatmap of adversarial example. The true label is `cinema' and the target label is `croquet ball'. The intensity of red indicates the level of importance assigned to each area, influencing the classifier prediction towards the target label.}
    \label{fig:cinema_croquetball}
\end{figure*}

Figs.~\ref{fig:black swan_weasel} and \ref{fig:cinema_croquetball} visualize the attention heatmap of the adversarial examples, with surrogate model ResNet-50 and target model ResNet-18. We  observe that the attention heatmaps w.r.t.~\name\ on the surrogate model are more similar to the heatmaps on target model, compared with other attack methods. This in part explains that the generated adversarial example of MCD  is more transferable to the target model.

\newpage
\section{Additional Experimental Results}
\label{sec:other_results}

\subsection{Result of masking the most important parameters}
\label{app:mask_most_parameter}

\input{tables/mmp}
Using the approximation in \cref{eqn:approx_importance}, we mask the most important parameters. The results are reported in \Cref{tab:mask_most_parameter}. As analyzed in \Cref{sec:method}, masking the most important parameters degrades the model’s attacking capability. A high masking ratio (e.g., 1 \%) causes the attack to fail, whereas a low ratio (e.g., 0.001 \%) leaves transferability unaffected, so the ASRs converge to the baseline RDI.

\subsection{Result of other related work}

We also include the empirical result of Ghost Networks  (GN) \cite{Ghost}. It  perturbs only
skip connections for residual networks and also uses one new model at each iteration. For a fair comparison, we  combine it with RDI and pick a larger iteration number for optimization. However, its performance is still much lower. 

\subsection{Ablation Study on DropConnect Probability}
\label{app:drop_prob}
\begin{figure}[ht] %
    \centering %
    \includegraphics[width=0.35\textwidth]{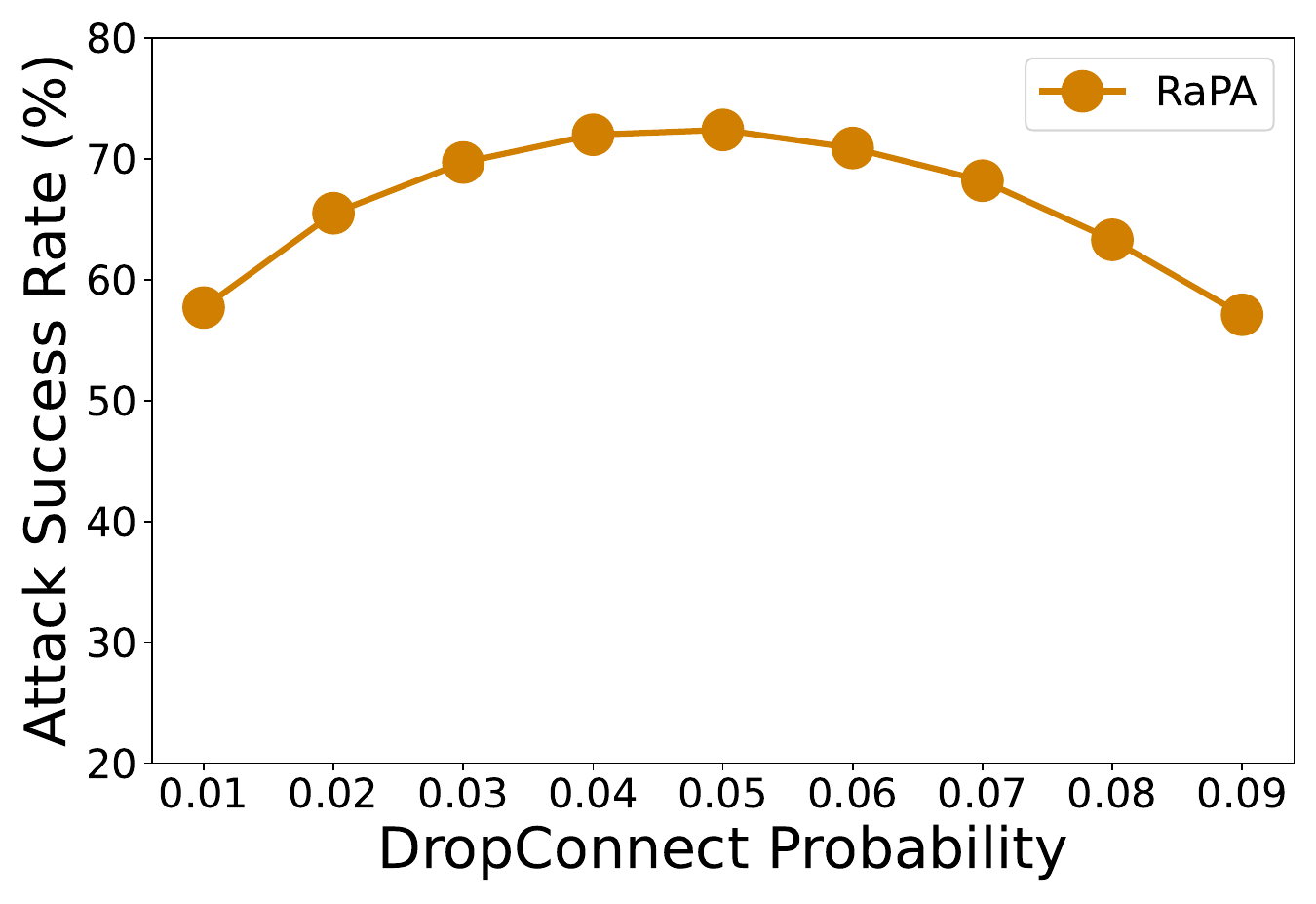} %
    \caption{ASRs(\%) averaged over 16 models with varying DropConnect probabilities, using ResNet-50 as surrogate .} %
    \label{fig:prob_ablation} %
\end{figure}

Experimental results of varying DropConnect probabilities are shown in \cref{fig:prob_ablation}, indicating a stable performance of \name\ across a range of different parameter values.

\subsection{Additional Attack Results}
\label{app:additionalresults}
\input{tables/other_cnn_results}

\input{tables/other_vit_results}

\paragraph{Evaluation on Same Architecture}
\cref{tab:other_cnn_results} and \cref{tab:other_vit_results} present additional attack results under the same-architecture transfer setting, following the configuration in \cref{sec:expsetting}. Our method \name\ consistently outperforms baselines by a large margin. Beyond these results, we further evaluate \name\ in more challenging scenarios, including various surrogate model architectures, multi-model ensembles, and untargeted attack settings.

\input{tables/modernmodel}
\paragraph{Evaluation on Newer and Larger Surrogate Models.}
To further validate the scalability of \name, we extend our evaluation to more recent and significantly larger surrogate models, including ConvNeXtV2-L \cite{convnextv2} (2023, 198M parameters), DINOv2-L \cite{dinov2} (2023, 300M), and the SSM-based MambaVision \cite{mambavision} (2025). As shown in \cref{tab:modern_models}, \name\ consistently outperforms the state-of-the-art CFM. Notably, the high source-to-source (Src$\to$Src) scores of CFM compared to \name\ further validate that CFM tends to overfit the surrogate model, whereas \name\ maintains better transferability.

\input{tables/imagenetv2}

\paragraph{Evaluation on Cross-Dataset Generalization.}
To verify the generalization of \name\ across different data distributions, we conduct additional experiments on the ImageNet-V2 dataset\cite{imagenetv2}. We evaluate the average ASRs across 10 CNN-based and 6 Transformer-based target models. As shown in \cref{tab:imagenetv2}, \name\ consistently outperforms all baseline methods. Notably, our method achieves a significant 16.4\% improvement in the challenging ViT-to-CNN transfer scenario. These results further validate that \name\ maintains its superior transferability with different evaluation dataset.

\paragraph{Evaluation on Multi-Model Ensembles.}
We also investigate the performance of \name\ in a multi-model ensemble setting. We employ an ensemble of five distinct source models (ResNet50, ResNet18, VGG16, DenseNet121, and Inception-v3) and test them against six Transformer-based target models. \name\ demonstrates superior performance in both single-model (45.0\% vs. 32.2\% for CFM) and multi-model ensemble settings (63.3\% vs. 60.7\% for CFM), proving its robustness across diverse source distributions.

\paragraph{Evaluation on Untargeted Attacks.}
While our primary focus is on the more challenging targeted attack task, we also evaluated \name\ in an untargeted setting. We observed that existing methods already achieve near-saturated performance; for instance, CFM reaches an average ASR of 98.2\% across 16 target models. In this regime, while \name\ yields similar ASRs, the performance gap between top-tier methods becomes statistically indistinguishable. Therefore, the targeted setting serves as a more effective benchmark for demonstrating the advantages of our approach.

\subsection{Model Diversity and Utility}
\label{sec:div_util}
\begin{figure}[t]
    \centering
    
        \centering
        \begin{subfigure}[b]{0.4\linewidth} 
            \centering
            \captionsetup{labelformat=offset}
            \includegraphics[width=\textwidth]{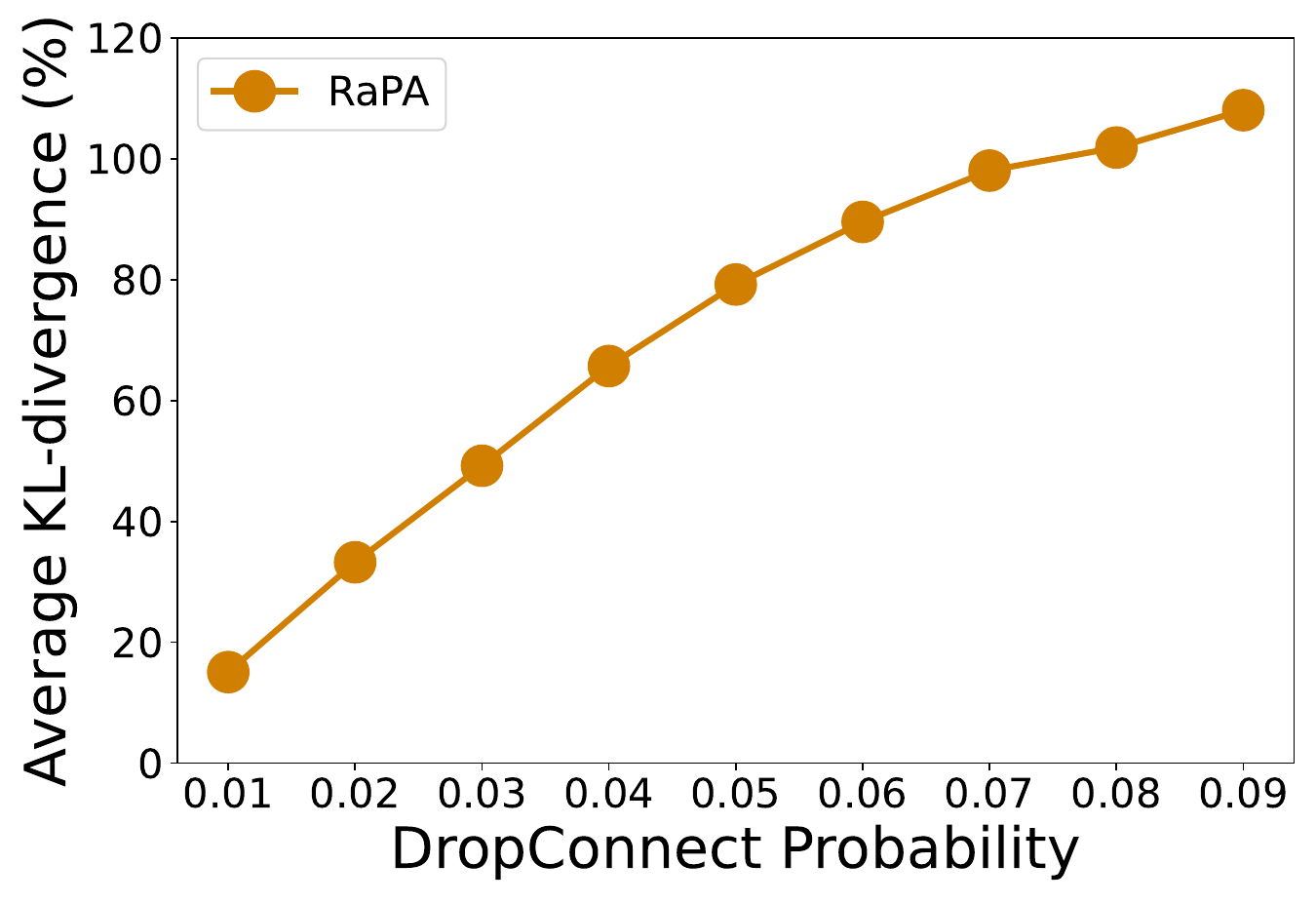}
            \caption{Diversity}
        \end{subfigure}
        \begin{subfigure}[b]{0.4\linewidth}
            \centering
            \captionsetup{labelformat=offset}
            \includegraphics[width=\linewidth]{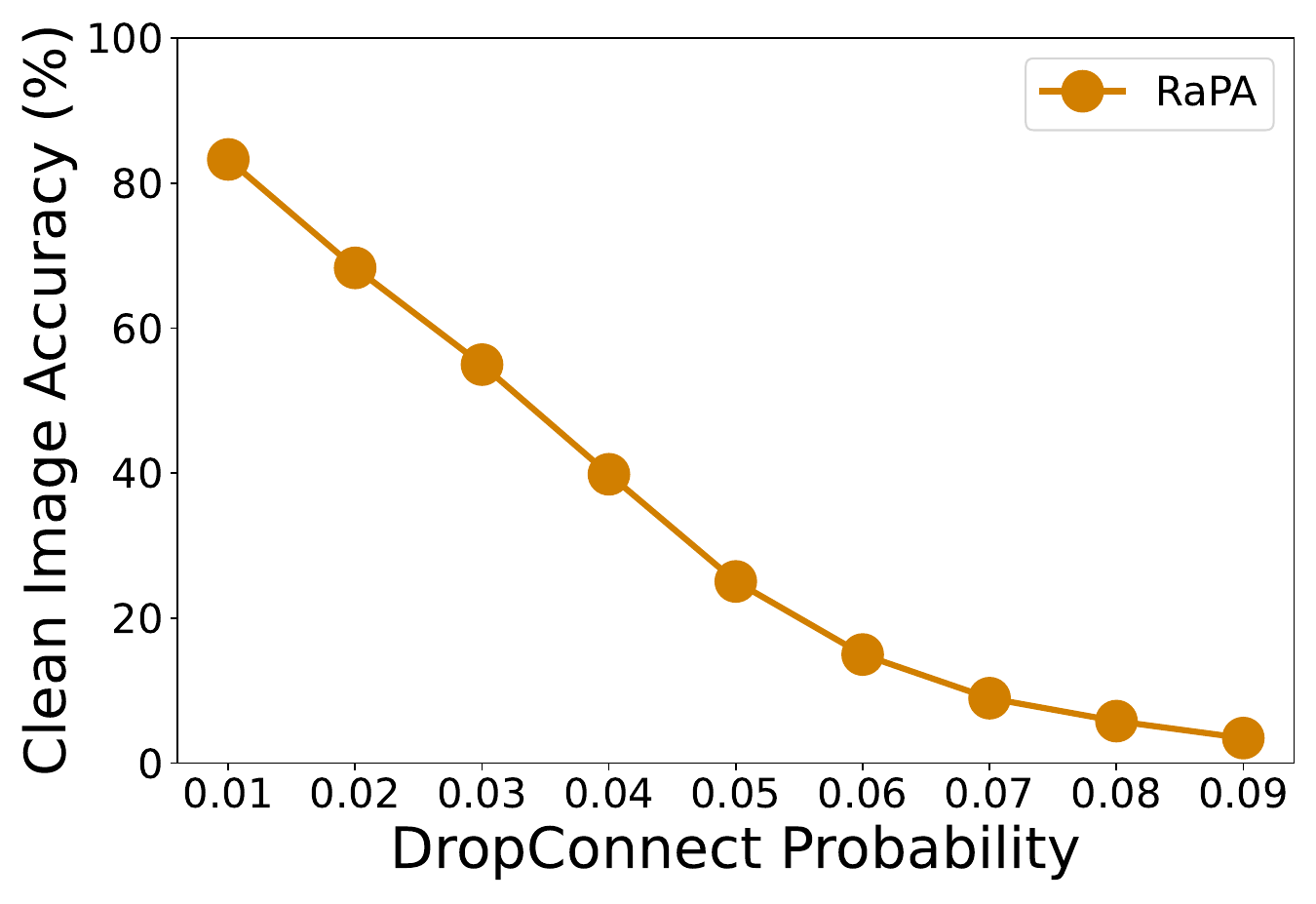}
            \caption{Utility}
        \end{subfigure}\hfill
    
    \caption{Diversity and Utility with increasing mask ratio.}
    \label{fig:div_util}
\end{figure}

We show the diversity and utility of constructed models. Here  diversity is measured by first computing the mean  KL-divergence between the output distributions of each pair of  model variants for an input and then averaging these values over the ImageNet-compatible dataset. Similarly, 
the utility is quantified by the mean top-1 accuracy over the model variants and the dataset. We use the same surrogate model ResNet-50 as in the ablation study. 

The results are presented in \Cref{fig:div_util}. As  DropConnect probability $p$ increases, the diversity among different variants improves while the utility decreases. As shown in \Cref{fig:prob_ablation} in the supplementary material, the average ASR using the same surrogate model increases with $p\in[0.01, 0.05]$  and then decreases with $p\in[0.05, 0.09]$, while the highest ASR is obtained at $p=0.05$. This verifies our discussion in \Cref{sec:discussion} that  the attack performance is affected by both diversity and utility. We  conclude that the better  performance of \name\ is due to the  improved diversification across the variants while keeping each variant sufficiently useful.

\newpage
\begin{center}
    {\LARGE \bf Ethical Statement}

\end{center}
While our work is intended for research and defense purposes, it could be misused by malicious actors to craft black-box adversarial attacks against safety-critical systems such as autonomous-driving or medical models. By demonstrating that these systems remain vulnerable even when attackers have no knowledge of their internal details, we aim to alert service providers and researchers to this threat and to accelerate the development of more robust deep-learning defenses.

%% file: tables/mmp.tex
\begin{table*}[ht]
\centering
\renewcommand{\arraystretch}{1.0}
\resizebox{0.75\textwidth}{!}{
\begin{tabular}{l c c c c c c c c c}
\toprule[0.15em]
\textbf{Method} & DI & GN & DWP & RDI & MMP(1\%) & MMP(0.1\%) & MMP(0.01\%) & MMP(0.001\%)  \\
\midrule
\textbf{ASR(\%)} & 26.8 & 29.7 & 48.0 & 42.0 & 0.2 & 27.7 & 36.7 & 38.9\\
\bottomrule[0.15em]
\end{tabular}%
}
\caption{Average ASRs when masking the most important parameters (MMP) in the surrogate model at mask ratios of 1 \%, 0.1 \%, 0.01 \%, and 0.001 \%. Results are the average over 16 target models  on the ImageNet-compatible dataset, with ResNet-50 as surrogate model. Detailed experimental setting can be found in \Cref{sec:expsetting}.}
\label{tab:mask_most_parameter}
\end{table*}

%% file: tables/other_cnn_results.tex
\begin{table*}[!hb]
\centering
\normalsize
\renewcommand{\arraystretch}{1.0}
\resizebox{0.9\textwidth}{!}{%
\begin{tabular}{lccccccccccc}
\toprule[0.15em]
\textbf{\begin{tabular}[c]{@{}l@{}}Source : RN50\end{tabular}} & \multicolumn{10}{c}{Target model} &  \\ \cmidrule(l){2-12}
Attack & RN18 & RN50 & VGG16 & Incv3 & EFB0 & DN121 & MBv2 & IRv2 & Incv4 & Xcep & Avg. \\ \midrule
DI & 60.1 & 98.7 & 64.6 & 8.1 & 28.6 & 77.3 & 28.4 & 13.5 & 21.1 & 15.1 & 41.5 \\
RDI & 80.6 & 98.7 & 75.4 & 32.8 & 52.6 & 87.4 & 49.6 & 47.8 & 51.1 & 40.7 & 61.7 \\
SI & 82.9 & \underline{98.8} & 73.7 & 60.8 & 66.0 & 90.0 & 59.4 & 66.9 & 68.7 & 60.6 & 72.8 \\
BSR & 87.9 & 98.5 & 84.1 & 54.9 & 70.0 & 91.6 & 65.3 & 66.5 & 68.9 & 63.3 & 75.1 \\
Admix & 83.2 & 98.1 & 78.1 & 57.1 & 67.8 & 86.8 & 70.3 & 61.9 & 64.4 & 64.5 & 73.2 \\
SIA & 84.5 & 98.3 & 79.3 & 39.0 & 67.1 & 88.4 & 70.1 & 46.0 & 53.4 & 50.9 & 67.7 \\
ODI & 77.5 & \textbf{98.9} & 83.0 & 63.9 & 69.9 & 89.8 & 61.2 & 71.0 & 71.1 & 69.2 & 75.5 \\
MUP & 90.3 & 97.7 & 85.6 & 61.8 & 75.0 & 91.6 & 79.3 & 70.9 &71.8 & 68.7 &79.3\\
CFM & \underline{92.0} & 98.0 & \underline{89.5} & \underline{74.8} & \underline{85.6} & \underline{93.4} & \underline{85.2} & \textbf{82.6} & \underline{82.5} & \underline{81.7} & \underline{86.5} \\
\rowcolor{Gray}
\name & \textbf{93.0} & 98.2 & \textbf{91.0} & \textbf{82.6} & \textbf{88.5} & \textbf{93.6} & \textbf{90.9} & \underline{82.1} & \textbf{85.5} & \textbf{85.0} & \textbf{89.0} \\
\midrule
\textbf{\begin{tabular}[c]{@{}l@{}}Source : DN121\end{tabular}} & \multicolumn{10}{c}{Target model} &  \\ \cmidrule(l){2-12}
Attack & RN18 & RN50 & VGG16 & Incv3 & EFB0 & DN121 & MBv2 & IRv2 & Incv4 & Xcep & Avg. \\ \midrule
DI & 28.1 & 36.4 & 35.8 & 6.5 & 14.5 & \textbf{99.0} & 10.1 & 9.8 & 13.9 & 10.1 & 26.4 \\
RDI & 50.9 & 54.4 & 46.7 & 19.3 & 26.9 & 98.6 & 20.6 & 29.8 & 30.5 & 24.7 & 40.2 \\
SI & 54.8 & 60.9 & 43.0 & 35.6 & 37.9 & 98.6 & 26.7 & 43.6 & 43.5 & 34.6 & 47.9 \\
BSR & 61.6 & 70.3 & 57.3 & 34.1 & 41.7 & 98.3 & 30.2 & 42.7 & 44.5 & 37.5 & 51.8 \\
Admix & 72.4 & 74.3 & 65.8 & 46.9 & 55.1 & 97.9 & 51.2 & 55.5 & 56.6 & 53.9 & 63.0 \\
SIA & 69.2 & 79.3 & 66.2 & 34.9 & 54.8 & 98.1 & 45.1 & 40.6 & 51.6 & 44.1 & 58.4 \\
ODI & 59.6 & 71.5 & 69.6 & 48.1 & 52.2 & \underline{98.9} & 35.5 & 58.1 & 60.0 & 52.7 & 60.6 \\
MUP & 79.9 & 86.9 & 75.2 & 50.3 & 60.5 & 97.7 & 51.4 & 65.7 & 66.5 & 61.0 & 69.5 \\
CFM & \underline{82.4} & \underline{88.9} & \underline{78.6} & \underline{60.9} & \underline{69.9} & 98.5 & \underline{60.5} & \underline{73.0} & \underline{72.7} & \underline{70.3} & \underline{75.6} \\
\rowcolor{Gray}
\name & \textbf{89.6} & \textbf{90.2} & \textbf{85.1} & \textbf{77.8} & \textbf{84.4} & 97.4 & \textbf{84.4} & \textbf{83.2} & \textbf{84.1} & \textbf{83.7} & \textbf{86.0} \\
\midrule
\textbf{\begin{tabular}[c]{@{}l@{}}Source : CLIP\end{tabular}} & \multicolumn{10}{c}{Target model} &  \\ \cmidrule(l){2-12}
Attack & RN18 & RN50 & VGG16 & Incv3 & EFB0 & DN121 & MBv2 & IRv2 & Incv4 & Xcep & Avg. \\ \midrule
DI & 0.1 & 0.1 & 0.1 & 0.0 & 0.0 & 0.1 & 0.2 & 0.0 & 0.0 & 0.1 & 0.1 \\
RDI & 0.5 & 0.4 & 0.2 & 0.4 & 0.9 & 0.4 & 0.3 & 0.4 & 0.2 & 0.7 & 0.4 \\
SI & 0.6 & 0.5 & 0.6 & 1.4 & 1.6 & 1.4 & 1.0 & 1.0 & 0.8 & 1.2 & 1.0 \\
BSR & 1.7 & 1.6 & 0.9 & 1.9 & 3.5 & 2.4 & 1.5 & 1.9 & 2.3 & 2.2 & 2.0 \\
Admix & 2.7 & 2.2 & 1.6 & 4.1 & 6.0 & 3.5 & 2.8 & 2.9 & 3.8 & 3.6 & 3.3 \\
SIA & 0.9 & 1.0 & 0.8 & 1.8 & 3.5 & 1.6 & 1.9 & 1.3 & 1.8 & 2.6 & 1.7 \\
ODI & \underline{7.7} & \underline{7.0} & \textbf{5.1} & \underline{11.0} & \underline{13.7} & \underline{13.0} & \underline{6.5} & \underline{11.3} & \underline{11.8} & \underline{11.5} & \underline{9.9} \\
CFM & 3.1 & 3.3 & 1.4 & 4.9 & 7.8 & 4.3 & 2.5 & 5.4 & 4.1 & 4.8 & 4.2 \\
\rowcolor{Gray}
\name & \textbf{8.8} & \textbf{9.5} & \underline{4.9} & \textbf{14.6} & \textbf{17.5} & \textbf{14.9} & \textbf{8.5} & \textbf{12.5} & \textbf{12.8} & \textbf{12.4} & \textbf{11.6} \\
\bottomrule[0.15em]
\end{tabular}%
}
\caption{Additional experimental results against ten target models on the ImageNet-Compatible dataset, using RN50, DN121 and CLIP as surrogate model, respectively. All the attack methods are combined with MI-TI. The best results are shown in bold and the second best results are underlined.}
\label{tab:other_cnn_results}
\vspace{-0.5cm}
\end{table*}

%% file: tables/other_vit_results.tex
\begin{table*}[!htbp]
\centering
\renewcommand{\arraystretch}{1.0}
\resizebox{0.6\textwidth}{!}{%
\begin{tabular}{lccccccc}
\toprule[0.15em]
\textbf{\begin{tabular}[c]{@{}l@{}}Source : Incv3\end{tabular}} & \multicolumn{6}{c}{Target model} &  \\ \cmidrule(l){2-8}
Attack & ViT & LeViT & ConViT & Twins & PiT & CLIP & Avg. \\ \midrule
DI & 0.1 & 0.8 & 0.1 & 0.3 & 0.5 & 0.2 & 0.3 \\
RDI & 0.1 & 3.1 & 0.6 & 2.0 & 1.9 & 0.0 & 1.3 \\
SI & 1.0 & 5.0 & 1.0 & 2.0 & 3.8 & 0.4 & 2.2 \\
SIA & 1.5 & 16.8 & 1.7 & 5.2 & 6.8 & 0.4 & 5.4 \\
BSR & 1.3 & 15.3 & 1.9 & 5.6 & 7.3 & 0.5 & 5.3 \\
Admix & 2.0 & 17.4 & 1.9 & 6.7 & 8.5 & 1.4 & 6.3 \\
ODI & 3.6 & 21.6 & 4.4 & 8.9 & 15.1 & 2.0 & 9.3 \\
MUP & 1.4&11.9&2.3&4.4&5.4&0.4&4.3 \\

CFM & \underline{9.7} & \underline{43.4} & \underline{8.8} & \underline{23.2} & \underline{26.4} & \underline{3.9} & \underline{19.2} \\
\rowcolor{Gray}
\name & \textbf{14.7} & \textbf{57.6} & \textbf{16.7} & \textbf{34.5} & \textbf{38.0} & \textbf{5.3} & \textbf{27.8} \\
\midrule
\textbf{\begin{tabular}[c]{@{}l@{}}Source : ViT\end{tabular}} & \multicolumn{6}{c}{Target model} &  \\ \cmidrule(l){2-8}
Attack & ViT & LeViT & ConViT & Twins & PiT & CLIP & Avg. \\ \midrule
DI & \textbf{100.0} & 11.3 & 15.6 & 7.4 & 14.8 & 3.2 & 25.4 \\
RDI & \textbf{100.0} & 31.5 & 38.9 & 23.6 & 36.2 & 7.7 & 39.6 \\
SI & \textbf{100.0} & 48.2 & 59.2 & 28.3 & 51.4 & 14.8 & 50.3 \\
SIA & \textbf{100.0} & 43.0 & 77.9 & 44.8 & 56.2 & 12.0 & 55.6 \\
BSR & \textbf{100.0} & 57.8 & 63.0 & 45.7 & 67.2 & 18.5 & 58.7 \\
Admix & 99.8 & 78.2 & 81.5 & 66.4 & 79.2 & 39.6 & 74.1 \\
SE &100 &65.6 & 75.7 & 56.4 & 72 & 20.4 & 65.0 \\
ODI & \textbf{100.0} & 74.8 & 68.3 & 63.1 & 81.7 & 31.6 & 69.9 \\
CFM & 99.9 & \underline{90.9} & \underline{94.2} & \underline{82.5} & \underline{91.6} & \underline{55.3} & \underline{85.7} \\
\rowcolor{Gray}
\name & 99.6 & \textbf{92.4} & \textbf{95.7} & \textbf{87.4} & \textbf{94.4} & \textbf{63.5} & \textbf{88.8} \\
\midrule
\textbf{\begin{tabular}[c]{@{}l@{}}Source : CLIP\end{tabular}} & \multicolumn{6}{c}{Target model} &  \\ \cmidrule(l){2-8}
Attack & ViT & LeViT & ConViT & Twins & PiT & CLIP & Avg. \\ \midrule
DI & 0.0 & 0.0 & 0.1 & 0.0 & 0.0 & \textbf{99.8} & 16.6 \\
RDI & 1.3 & 1.5 & 0.6 & 0.4 & 1.0 & \textbf{99.8} & 17.4 \\
SI & 3.1 & 4.5 & 2.5 & 1.7 & 6.2 & 99.6 & 19.6 \\
SIA & 7.7 & 7.9 & 5.3 & 3.6 & 8.3 & 99.7 & 22.1 \\
BSR & 7.0 & 9.2 & 4.9 & 2.9 & 8.5 & 99.5 & 22.0 \\
Admix & 7.3 & 11.9 & 5.5 & 4.2 & 11.1 & 99.0 & 23.2 \\
ODI & \underline{19.1} & \underline{24.3} & \underline{13.1} & \underline{11.7} & \underline{22.7} & \textbf{99.8} & \underline{31.8} \\
CFM & 13.3 & 15.4 & 8.6 & 4.7 & 13.5 & \textbf{99.8} & 25.9 \\
\rowcolor{Gray}
\name & \textbf{32.5} & \textbf{35.5} & \textbf{25.1} & \textbf{16.7} & \textbf{33.0} & 98.4 & \textbf{40.2} \\
\bottomrule[0.15em]
\end{tabular}%
}
\caption{Additional experimental results against five transformer-based target models on the ImageNet-Compatible dataset, using RN50, DN121 and CLIP as surrogate model, respectively. All methods are combined with MI-TI. The best results are shown in bold, and the second best results are underlined.}
\label{tab:other_vit_results}
\end{table*}

%% file: tables/modernmodel.tex
\begin{table}[h]
\centering
\begin{tabular}{l|l|c|c|c}
\hline
\textbf{Source Model} & \textbf{Method} & \textbf{CNNs Avg} & \textbf{ViTs Avg} & \textbf{Src$\to$Src} \\
\hline
\multirow{2}{*}{\begin{tabular}[c]{@{}l@{}}MambaVision-T \end{tabular}} & CFM & 12.5 & 40.2 & 91.9 \\
 & \textbf{RaPA} & \textbf{63.8} & \textbf{64.9} & 89.7 \\
\hline
\multirow{2}{*}{\begin{tabular}[c]{@{}l@{}}DINOv2-Large\end{tabular}} & CFM & 20.0 & 27.7 & 100.0 \\
 & \textbf{RaPA} & \textbf{41.4} & \textbf{49.3} & 98.7 \\
\hline
\multirow{2}{*}{\begin{tabular}[c]{@{}l@{}}ConvNeXtV2-L \end{tabular}} & CFM & 20.6 & 28.4 & 100.0 \\
 & \textbf{RaPA} & \textbf{46.4} & \textbf{56.7} & 100.0 \\
\hline
\end{tabular}
\caption{Comparisons on newer and larger source models. ASR (\%) is reported. Src$\to$Src prove that CFM overfit to surrogate model}
\label{tab:modern_models}
\end{table}

%% file: tables/imagenetv2.tex
\begin{table}[h]
    \centering
    \begin{tabular}{l|cc|cc}
        \hline
        \multirow{2}{*}{\textbf{Method}} & \multicolumn{2}{c|}{\textbf{Source: RN50}} & \multicolumn{2}{c}{\textbf{Source: ViT}} \\
         & \textbf{CNN} & \textbf{ViT} & \textbf{CNN} & \textbf{ViT} \\
        \hline
        CFM & 81.9 & 52.9 & 56.6 & 86.9 \\
        FTM & 83.2 & 54.7 & 20.3 & 35.6 \\
        \textbf{RaPA (Ours)} & \textbf{86.3} & \textbf{58.7} & \textbf{73.0} & \textbf{91.0} \\
        \hline
    \end{tabular}
    \caption{ASR (\%) on ImageNet-V2. RaPA vs. baselines (CFM, FTM) using RN50 and ViT surrogates transferred to 10 CNN and 6 Transformer models.}
      \label{tab:imagenetv2}
\end{table}

%% file: main.bib
@String(NIPS= {Adv. Neural Inform. Process. Syst.})

@String(AAAI = {AAAI})

@String(NIPS  = {NeurIPS})

@article{TransferSurvey,
  title={A Survey on Transferability of Adversarial Examples Across Deep Neural Networks},
  author={Gu, Jindong and Jia, Xiaojun and de Jorge, Pau and Yu, Wenqian and Liu, Xinwei and Ma, Avery and Xun, Yuan and Hu, Anjun and Khakzar, Ashkan and Li, Zhijiang and others},
  journal={Transactions on Machine Learning Research},
  year={2024}
}

@inproceedings{FGSM,
  title={{Explaining and harnessing adversarial examples}},
  author={Goodfellow, Ian J and Shlens, Jonathon and Szegedy, Christian},
  booktitle={IEEE/CVF Conference on Computer Vision and Pattern Recognition},
  year={2015}
}

@incollection{I-FGSM,
  title={Adversarial examples in the physical world},
  author={Kurakin, Alexey and Goodfellow, Ian J and Bengio, Samy},
  booktitle={Artificial Intelligence Safety and Security},
  pages={99--112},
  year={2018},
  publisher={Chapman and Hall/CRC}
}

@inproceedings{DI,
  title={Improving transferability of adversarial examples with input diversity},
  author={Xie, Cihang and Zhang, Zhishuai and Zhou, Yuyin and Bai, Song and Wang, Jianyu and Ren, Zhou and Yuille, Alan L},
  booktitle={IEEE/CVF Conference on Computer Vision and Pattern Recognition},
  year={2019}
}

@inproceedings{TI,
  title={Evading defenses to transferable adversarial examples by translation-invariant attacks},
  author={Dong, Yinpeng and Pang, Tianyu and Su, Hang and Zhu, Jun},
  booktitle={IEEE/CVF Conference on Computer Vision and Pattern Recognition},
  year={2019}
}

@inproceedings{RDI,
  title={Improving the transferability of adversarial examples with resized-diverse-inputs, diversity-ensemble and region fitting},
  author={Zou, Junhua and Pan, Zhisong and Qiu, Junyang and Liu, Xin and Rui, Ting and Li, Wei},
  booktitle={European Conference on Computer Vision},
  year={2020},
}

@inproceedings{Admix,
  title={Admix: Enhancing the transferability of adversarial attacks},
  author={Wang, Xiaosen and He, Xuanran and Wang, Jingdong and He, Kun},
  booktitle={International Conference on Computer Vision},
  year={2021}
}

@inproceedings{ODI,
  title={Improving the transferability of targeted adversarial examples through object-based diverse input},
  author={Byun, Junyoung and Cho, Seungju and Kwon, Myung-Joon and Kim, Hee-Seon and Kim, Changick},
  booktitle={IEEE/CVF Conference on Computer Vision and Pattern Recognition},
  year={2022}
}

@inproceedings{CFM,
  title={Introducing competition to boost the transferability of targeted adversarial examples through clean feature mixup},
  author={Byun, Junyoung and Kwon, Myung-Joon and Cho, Seungju and Kim, Yoonji and Kim, Changick},
  booktitle={IEEE/CVF Conference on Computer Vision and Pattern Recognition},
  year={2023}
}

@inproceedings{BSR,
  title={Boosting adversarial transferability by block shuffle and rotation},
  author={Wang, Kunyu and He, Xuanran and Wang, Wenxuan and Wang, Xiaosen},
  booktitle={IEEE/CVF Conference on Computer Vision and Pattern Recognition},
  year={2024}
}

@inproceedings{MI,
  title={Boosting adversarial attacks with momentum},
  author={Dong, Yinpeng and Liao, Fangzhou and Pang, Tianyu and Su, Hang and Zhu, Jun and Hu, Xiaolin and Li, Jianguo},
  booktitle={IEEE/CVF Conference on Computer Vision and Pattern Recognition},
  year={2018}
}

@inproceedings{SI,
  title={{Nesterov Accelerated Gradient and Scale Invariance for Adversarial Attacks}},
  author={Jiadong Lin and Chuanbiao Song and Kun He and Liwei Wang and John E. Hopcroft},
  booktitle={International Conference on Learning Representations},
  year={2020}
}

@inproceedings{VT,
  title={Enhancing the transferability of adversarial attacks through variance tuning},
  author={Wang, Xiaosen and He, Kun},
  booktitle={IEEE/CVF Conference on Computer Vision and Pattern Recognition},
  year={2021}
}

@inproceedings{Po+trip,
  author = {Li, Maosen and Deng, Cheng and Li, Tengjiao and Yan, Junchi and Gao, Xinbo and Huang, Heng},
  title = {Towards Transferable Targeted Attack},
  booktitle={IEEE/CVF Conference on Computer Vision and Pattern Recognition},
  month = {June},
  year = {2020}
}

@inproceedings{Logit,
  title={On success and simplicity: A second look at transferable targeted attacks},
  author={Zhao, Zhengyu and Liu, Zhuoran and Larson, Martha},
  booktitle ={Advances in Neural Information Processing Systems},
  volume={34},
  year={2021}
}

@inproceedings{R&P,
  title={{Mitigating adversarial effects through randomization}},
  author={Xie, Cihang and Wang, Jianyu and Zhang, Zhishuai and Ren, Zhou and Yuille, Alan},
  booktitle={International Conference on Learning Representations},
  year={2018}
}

@inproceedings{JPEGandBit,
  title={{Countering adversarial images using input transformations}},
  author={Guo, Chuan and Rana, Mayank and Cisse, Moustapha and Maaten, Laurens Van Der},
  booktitle={International Conference on Learning Representations},
  year={2018}
}

@inproceedings{HGD,
  title={Defense against adversarial attacks using high-level representation guided denoiser},
  author={Liao, Fangzhou and Liang, Ming and Dong, Yinpeng and Pang, Tianyu and Hu, Xiaolin and Zhu, Jun},
  booktitle={IEEE/CVF Conference on Computer Vision and Pattern Recognition},
  year={2018}
}

@inproceedings{DiffPure,
  title={Diffusion Models for Adversarial Purification},
  author={Nie, Weili and Guo, Brandon and Huang, Yujia and Xiao, Chaowei and Vahdat, Arash and Anandkumar, Anima},
  booktitle = {International Conference on Machine Learning (ICML)},
  year={2022}
}

@inproceedings{First,
  author={Christian Szegedy and Wojciech Zaremba and Ilya Sutskever and Joan Bruna and Dumitru Erhan and Ian Goodfellow and Rob Fergus},
  title={Intriguing properties of neural networks},
  booktitle={International Conference on Learning Representations},
  year=2014
}

@inproceedings{Attention,
  title={Attention is all you need},
  author={Vaswani, Ashish and Shazeer, Noam and Parmar, Niki and Uszkoreit, Jakob and Jones, Llion and Gomez, Aidan N and Kaiser, {\L}ukasz and Polosukhin, Illia},
  booktitle={Advances in Neural Information Processing Systems},
  volume={30},
  year={2017}
}

@inproceedings{BN,
  title={Batch normalization: Accelerating deep network training by reducing internal covariate shift},
  author={Ioffe, Sergey and Szegedy, Christian},
  booktitle={International Conference on Machine Learning},
  year={2015},
}

@inproceedings{VGG16,
  title={Very deep convolutional networks for large-scale image recognition},
  author={Simonyan, Karen and Zisserman, Andrew},
  booktitle={International Conference on Learning Representations},
  year={2015},
}

@inproceedings{ResNet,
  title={Deep residual learning for image recognition},
  author={He, Kaiming and Zhang, Xiangyu and Ren, Shaoqing and Sun, Jian},
  booktitle={IEEE/CVF Conference on Computer Vision and Pattern Recognition},
  year={2016}
}

@inproceedings{DenseNet,
  title={Densely connected convolutional networks},
  author={Huang, Gao and Liu, Zhuang and Van Der Maaten, Laurens and Weinberger, Kilian Q},
  booktitle={IEEE/CVF Conference on Computer Vision and Pattern Recognition},
  year={2017}
}

@inproceedings{Xception,
  title={Xception: Deep learning with depthwise separable convolutions},
  author={Chollet, François},
  booktitle={IEEE/CVF Conference on Computer Vision and Pattern Recognition},
  year={2017}
}

@inproceedings{Mobilenetv2,
  title={Mobilenetv2: Inverted residuals and linear bottlenecks},
  author={Sandler, Mark and Howard, Andrew and Zhu, Menglong and Zhmoginov, Andrey and Chen, Liang-Chieh},
  booktitle={IEEE/CVF Conference on Computer Vision and Pattern Recognition},
  year={2018}
}

@inproceedings{Efficientnet,
  title={Efficientnet: Rethinking model scaling for convolutional neural networks},
  author={Tan, Mingxing and Le, Quoc},
  booktitle={International Conference on Machine Learning},
  year={2019}
}

@inproceedings{Inception-v4,
  title={Inception-v4, inception-resnet and the impact of residual connections on learning},
  author={Szegedy, Christian and Ioffe, Sergey and Vanhoucke, Vincent and Alemi, Alexander},
  booktitle={AAAI Conference on Artificial Intelligence},
  year={2017}
}

@inproceedings{Inception-v3,
  title={Rethinking the inception architecture for computer vision},
  author={Szegedy, Christian and Vanhoucke, Vincent and Ioffe, Sergey and Shlens, Jon and Wojna, Zbigniew},
  booktitle={IEEE/CVF Conference on Computer Vision and Pattern Recognition},
  year={2016}
}

@inproceedings{ViT,
  title={An Image is Worth 16x16 Words: Transformers for Image Recognition at Scale},
  author={Dosovitskiy, Alexey and Beyer, Lucas and Kolesnikov, Alexander and Weissenborn, Dirk and Zhai, Xiaohua and Unterthiner, Thomas and Dehghani, Mostafa and Minderer, Matthias and Heigold, G and Gelly, S and others},
  booktitle={International Conference on Learning Representations},
  year={2020}
}

@inproceedings{LeViT,
  title={Levit: a vision transformer in convnet's clothing for faster inference},
  author={Graham, Benjamin and El-Nouby, Alaaeldin and Touvron, Hugo and Stock, Pierre and Joulin, Armand and Jégou, Hervé and Douze, Matthijs},
  booktitle={International Conference on Computer Vision},
  year={2021}
}

@inproceedings{ConViT,
  title={Convit: Improving vision transformers with soft convolutional inductive biases},
  author={d’Ascoli, Stéphane and Touvron, Hugo and Leavitt, Matthew L and Morcos, Ari S and Biroli, Giulio and Sagun, Levent},
  booktitle={International Conference on Machine Learning},
  year={2021}
}

@inproceedings{Twins,
  title={Twins: Revisiting the design of spatial attention in vision transformers},
  author={Chu, Xiangxiang and Tian, Zhi and Wang, Yuqing and Zhang, Bo and Ren, Haibing and Wei, Xiaolin and Xia, Huaxia and Shen, Chunhua},
  booktitle={Advances in Neural Information Processing Systems},
  year={2021}
}

@inproceedings{PiT,
  title={Rethinking spatial dimensions of vision transformers},
  author={Heo, Byeongho and Yun, Sangdoo and Han, Dongyoon and Chun, Sanghyuk and Choe, Junsuk and Oh, Seong Joon},
  booktitle={International Conference on Computer Vision},
  year={2021}
}

@inproceedings{CLIP,
  title={Learning transferable visual models from natural language supervision},
  author={Radford, Alec and Kim, Jong Wook and Hallacy, Chris and Ramesh, Aditya and Goh, Gabriel and Agarwal, Sandhini and Sastry, Girish and Askell, Amanda and Mishkin, Pamela and Clark, Jack and others},
  booktitle={International Conference on Machine Learning},
  year={2021}
}

@inproceedings{ImageNet,
  title={Imagenet: A large-scale hierarchical image database},
  author={Deng, Jia and Dong, Wei and Socher, Richard and Li, Li-Jia and Li, Kai and Fei-Fei, Li},
  booktitle={IEEE/CVF Conference on Computer Vision and Pattern Recognition},
  year={2009}
}

@inproceedings{advResNet,
  title={Do adversarially robust imagenet models transfer better?},
  author={Salman, Hadi and Ilyas, Andrew and Engstrom, Logan and Kapoor, Ashish and Madry, Aleksander},
  booktitle={Advances in Neural Information Processing Systems},
  year={2020}
}

@inproceedings{advIR,
  title={Adversarial attacks and defences competition},
  author={Kurakin, Alexey and Goodfellow, Ian and Bengio, Samy and Dong, Yinpeng and Liao, Fangzhou and Liang, Ming and Pang, Tianyu and Zhu, Jun and Hu, Xiaolin and Xie, Cihang and others},
  booktitle={The NIPS'17 Competition: Building Intelligent Systems},
  year={2018},
}

@inproceedings{mahmood2021robustness,
  title={On the robustness of vision transformers to adversarial examples},
  author={Mahmood, Kaleel and Mahmood, Rigel and Van Dijk, Marten},
  booktitle={International Conference on Computer Vision},
  year={2021}
}

@inproceedings{DropConnect,
  title={Regularization of neural networks using dropconnect},
  author={Wan, Li and Zeiler, Matthew and Zhang, Sixin and Le Cun, Yann and Fergus, Rob},
  booktitle={International Conference on Machine Learning},
  year={2013}
}

@inproceedings{T-sea,
  title={T-sea: Transfer-based self-ensemble attack on object detection},
  author={Huang, Hao and Chen, Ziyan and Chen, Huanran and Wang, Yongtao and Zhang, Kevin},
  booktitle={IEEE/CVF Conference on Computer Vision and Pattern Recognition},
  year={2023}
}

@inproceedings{MUP,
  author={Yang, Dingcheng and Yu, Wenjian and Xiao, Zihao and Luo, Jiaqi},
  booktitle={International Joint Conference on Neural Networks}, 
  title={Generating Adversarial Examples with Better Transferability via Masking Unimportant Parameters of Surrogate Model}, 
  year={2023},     
}

@inproceedings{Ghost,
  title={Learning transferable adversarial examples via ghost networks},
  author={Li, Yingwei and Bai, Song and Zhou, Yuyin and Xie, Cihang and Zhang, Zhishuai and Yuille, Alan},
  booktitle={AAAI Conference on Artificial Intelligence},
  year={2020}
}

@inproceedings{SETR,
  title={On improving adversarial transferability of vision transformers},
  author={Naseer, Muzammal and Ranasinghe, Kanchana and Khan, Salman and Khan, Fahad Shahbaz and Porikli, Fatih},
  booktitle={International Conference on Learning Representations},
  year={2022}
}

@inproceedings{Liu2016DelvingIT,
  title={Delving into Transferable Adversarial Examples and Black-box Attacks},
  author={Yanpei Liu and Xinyun Chen and Chang Liu and Dawn Xiaodong Song},
 booktitle={International Conference on Learning Representations},
  year={2017}
}

@inproceedings{TRS,
  title={TRS: Transferability Reduced Ensemble via Encouraging Gradient Diversity and Model Smoothness},
  author={ Yang, Zhuolin  and  Li, Linyi  and  Xu, Xiaojun  and  Zuo, Shiliang  and  Chen, Qian  and  Rubinstein, Benjamin  and  Zhang, Ce  and  Li, Bo },
booktitle={Advances in Neural Information Processing Systems},
  year={2021},
}

@inproceedings{Chen2023RethinkingME,
  title={Rethinking Model Ensemble in Transfer-based Adversarial Attacks},
  author={Huanran Chen and Yichi Zhang and Yinpeng Dong and Junyi Zhu},
 booktitle={International Conference on Learning Representations},
  year={2024}
}

@article{liu2024scaling,
  title={Scaling Laws for Black box Adversarial Attacks},
  author={Liu, Chuan and Chen, Huanran and Zhang, Yichi and Dong, Yinpeng and Zhu, Jun},
  journal={arXiv preprint arXiv:2411.16782},
  year={2024}
}

@article{lecun1989optimal,
  title={Optimal brain damage},
  author={LeCun, Yann and Denker, John and Solla, Sara},
  journal={Advances in neural information processing systems},
  volume={2},
  year={1989}
}

@inproceedings{fang2023depgraph,
  title={Depgraph: Towards any structural pruning},
  author={Fang, Gongfan and Ma, Xinyin and Song, Mingli and Mi, Michael Bi and Wang, Xinchao},
  booktitle={IEEE/CVF Conference on Computer Vision and Pattern Recognition},
  year={2023}
}

@inproceedings{molchanov2019importance,
  title={Importance estimation for neural network pruning},
  author={Molchanov, Pavlo and Mallya, Arun and Tyree, Stephen and Frosio, Iuri and Kautz, Jan},
  booktitle={IEEE/CVF conference on computer vision and pattern recognition},
  year={2019}
}

@inproceedings{SIA,
  title={Structure invariant transformation for better adversarial transferability},
  author={Wang, Xiaosen and Zhang, Zeliang and Zhang, Jianping},
  booktitle={Proceedings of the IEEE/CVF International Conference on Computer Vision},
  year={2023}
}

@inproceedings{SASD-WS,
  title={Improving transferable targeted adversarial attacks with model self-enhancement},
  author={Wu, Han and Ou, Guanyan and Wu, Weibin and Zheng, Zibin},
  booktitle={Proceedings of the IEEE/CVF Conference on Computer Vision and Pattern Recognition},
  year={2024}
}

@inproceedings{DWP,
  title={Enhancing Targeted Attack Transferability via Diversified Weight Pruning},
  author={Wang, Hung-Jui and Wu, Yu-Yu and Chen, Shang-Tse},
  booktitle={Proceedings of the IEEE/CVF Conference on Computer Vision and Pattern Recognition Workshops},
  year={2024},
}

@inproceedings{FTM,
  title={Improving Transferable Targeted Attacks with Feature Tuning Mixup},
  author={Liang, Kaisheng and Dai, Xuelong and Li, Yanjie and Wang, Dong and Xiao, Bin},
  booktitle={Proceedings of the IEEE/CVF Conference on Computer Vision and Pattern Recognition},
  pages={25802--25811},
  year={2025}
}

@inproceedings{DSM,
  title={Boosting the adversarial transferability of surrogate models with dark knowledge},
  author={Yang, Dingcheng and Xiao, Zihao and Yu, Wenjian},
  booktitle={2023 IEEE 35th International Conference on Tools with Artificial Intelligence},
  year={2023},
}

@incollection{nips_dataset,
  title={Adversarial attacks and defences competition},
  author={Kurakin, Alexey and Goodfellow, Ian and Bengio, Samy and Dong, Yinpeng and Liao, Fangzhou and Liang, Ming and Pang, Tianyu and Zhu, Jun and Hu, Xiaolin and Xie, Cihang and others},
  booktitle={The NIPS'17 Competition: Building Intelligent Systems},
  pages={195--231},
  year={2018},
  publisher={Springer}
}

@inproceedings{imagenetv2,
  title={Do imagenet classifiers generalize to imagenet?},
  author={Recht, Benjamin and Roelofs, Rebecca and Schmidt, Ludwig and Shankar, Vaishaal},
  booktitle={International conference on machine learning},
  pages={5389--5400},
  year={2019},
  organization={PMLR}
}

@article{convnextv2,
  title={ConvNeXt V2: Co-designing and Scaling ConvNets with Masked Autoencoders},
  author={Sanghyun Woo and Shoubhik Debnath and Ronghang Hu and Xinlei Chen and  Zhuang Liu and In So Kweon and Saining Xie},
  year={2023},
  journal={arXiv preprint arXiv:2301.00808},
}

@inproceedings{mambavision,
  title={Mambavision: A hybrid mamba-transformer vision backbone},
  author={Hatamizadeh, Ali and Kautz, Jan},
  booktitle={Proceedings of the IEEE/CVF Conference on Computer Vision and Pattern Recognition},
  year={2025}
}

@misc{dinov2,
  title={DINOv2: Learning Robust Visual Features without Supervision},
  author={Oquab, Maxime and Darcet, Timothée and Moutakanni, Theo and Vo, Huy V. and Szafraniec, Marc and Khalidov, Vasil and Fernandez, Pierre and Haziza, Daniel and Massa, Francisco and El-Nouby, Alaaeldin and Howes, Russell and Huang, Po-Yao and Xu, Hu and Sharma, Vasu and Li, Shang-Wen and Galuba, Wojciech and Rabbat, Mike and Assran, Mido and Ballas, Nicolas and Synnaeve, Gabriel and Misra, Ishan and Jegou, Herve and Mairal, Julien and Labatut, Patrick and Joulin, Armand and Bojanowski, Piotr},
  journal={arXiv:2304.07193},
  year={2023}
}
